\documentclass[msom,nonblindrev]{informsnohead}
\usepackage[top=1.5cm, bottom=3.5cm, left=1.5cm, right=3cm]{geometry}

\RequirePackage{endnotes}

\usepackage{graphicx} % Allows including images
\usepackage{booktabs} % Allows the use of \toprule, \midrule and \bottomrule in tables
\usepackage{subfigure, epsfig}
\usepackage{natbib}
\usepackage{bm}
\usepackage{multicol}
\usepackage{multirow}
\usepackage{amsmath}
\usepackage{mathrsfs}
\usepackage{caption}
\usepackage[section]{placeins}
\usepackage{float}
\usepackage[ruled,linesnumbered]{algorithm2e}
\usepackage{subfigure}
\usepackage[colorlinks=TRUE,citecolor=MyDarkBlue,urlcolor=MyDarkBlue,linkcolor=MyDarkBlue]{hyperref}
\definecolor{MyDarkBlue}{RGB}{158,0,0}
\usepackage{threeparttable}

\OneAndAHalfSpacedXI
\SetKwInput{KwIn}{Input}
\SetKwInput{KwOut}{Output}
\SetKw{KwRet}{Return}

 \bibpunct[, ]{(}{)}{,}{a}{}{,}%
 \def\bibfont{\small}%
\TheoremsNumberedThrough     % Preferred (Theorem 1, Lemma 1, Theorem 2)
\ECRepeatTheorems

\EquationsNumberedThrough    % Default: (1), (2), ...
\begin{document}
%%%%%%%%%%%%%%%%

\RUNAUTHOR{Liao et al.}

\RUNTITLE{DL for Perishables with Human Knowledge}

\TITLE{Deep Learning for Perishable Inventory Systems with Human Knowledge}

\ARTICLEAUTHORS{%
\AUTHOR{Xuan Liao}
\AFF{Shanghai Jiao Tong University, \EMAIL{
xuanliao@sjtu.edu.cn}} %, \URL{}}
\AUTHOR{Zhenkang Peng}
\AFF{The Chinese University of Hong Kong,  \EMAIL{zhenkang.peng@cuhk.edu.hk}}
\AUTHOR{Ying Rong}
\AFF{Shanghai Jiao Tong University,  \EMAIL{yrong@sjtu.edu.cn}}
} % end of the block

\ABSTRACT{Managing perishable products with limited lifetimes is a fundamental challenge in inventory management, as poor ordering decisions can quickly lead to stockouts or excessive waste. We study a perishable inventory system with random lead times in which both the demand process and the lead time distribution are unknown to the decision maker. We focus on a practical setting where the decision maker must place orders using limited historical data together with observed covariates and current system states. To improve learning efficiency under limited historical data, we adopt a well-established marginal cost accounting scheme that, even in this dynamic setting, assigns each order a single lifetime cost and yields a unified loss function for end-to-end learning. This supports training a deep learning–based policy that maps observed covariates and system states directly to an order quantity. We develop two end-to-end variants: a purely black-box approach that outputs the order quantity directly (E2E-BB), and a structure-guided approach that embeds the projected inventory level (PIL) policy, capturing the impact of inventory status through an explicit computation rather than additional learning (E2E-PIL). We further show that the objective induced by E2E-PIL is homogeneous of degree one, enabling a boosting technique from operational data analytics (ODA) that yields an enhanced policy (E2E-BPIL). Experiments on both synthetic and real data establish a robust performance ordering: E2E-BB is dominated by E2E-PIL, which is further improved by E2E-BPIL. Using an excess-risk decomposition, we show that embedding heuristic policy structure, while not fully optimal, can substantially reduce effective model complexity and improve learning efficiency with only a modest loss of flexibility. More broadly, our results suggest a general principle for operations management: deep learning based decision tools become more effective and robust when guided by human knowledge, highlighting the value of integrating advanced analytics with established inventory theory.

% Enter your abstract
}%


\maketitle
%%%%%%%%%%%%%%%%%%%%%%%%%%%%%%%%%%%%%%%%%%%%%%%%%%%%%%%%%%%%%%%%%%%%%%

\section{Introduction}

Inventory management is central to operational performance because inventory both ties up working capital and entails a range of costs. Practitioner benchmarks suggest that annual inventory carrying costs can reach roughly 20–30\% of inventory value, and that improvements in inventory practices can reduce excess inventory by as much as 25\%.\footnote{\href{https://wifitalents.com/inventory-statistics/}{https://wifitalents.com/inventory-statistics/}
, accessed on September 1, 2025.} These magnitudes underscore the economic stakes of replenishment decisions and have motivated an extensive literature in operations management, from the classical Newsvendor model \citep{scarf1958min} to multi-period stochastic control formulations and their many extensions \citep{zipkin2000foundations}. A defining strength of this literature is its ability to derive structured, interpretable policies. At the same time, such tractability and structural clarity often hinge on modeling assumptions, and small changes in these assumptions can fundamentally alter the structure of optimal policies. For example, with backorder demand, base-stock policies are optimal in broad classes of multi-period inventory systems with positive lead times \citep{scarf1958inventory}; under lost sales, this optimality generally breaks down, and the structure of optimal policies is not known in general \citep{ChenJiangZhangZhou2024}. This sensitivity highlights a recurring tension: assumption-driven theory yields elegant structure, but the resulting policies may not transfer cleanly to richer operational settings. Nevertheless, concepts such as base-stock and $(s,S)$ policies remain widely used in practice, often supplemented by ad-hoc customization to accommodate operational constraints and data realities.

This gap between stylized structure and operational complexity has spurred interest in data-driven decision making. In particular, over the past several years, the rapid maturation of deep learning (DL) has fueled growing interest in end-to-end (E2E) learning for inventory control. While DL was originally developed and popularized as a flexible paradigm for prediction \citep{lecun2015deep}, it has increasingly been adapted to prescribe decisions by learning mappings from observed system information to actions through deep neural networks (DNNs). A growing body of work applies such approaches to inventory settings \citep[e.g.,][]{oroojlooyjadid2020applying,han2023deep,qi2023practical}, with emerging evidence of industrial adoption (e.g., \citet{qi2023practical} described deployment at JD.com). Because these methods can perform well even when the structural properties of optimal policies are unknown or difficult to exploit, their popularity raises a natural concern: Can black-box E2E learning supplant the structured insights and policy forms developed over decades of inventory theory? We revisit this question and ask more specifically: \textit{Does human knowledge, particularly policy structure distilled from inventory theory, still add value when powerful E2E models can be trained from data?}

If black-box E2E models truly made inventory theory obsolete, they should perform reliably in the settings where theory is hardest to apply: high-dimensional stochastic systems in which demand and lead time primitives are unknown, potentially correlated, and nonstationary. We therefore focus on a multi-period perishable inventory system with random lead times, a setting that is both practically important and theoretically challenging. In practice, perishable operations have attracted growing attention \citep[e.g.,][]{li2022separation,atan2025displaying} and its importance is reflected in market size and growth: the global perishable prepared food segment is projected to increase from \$103.99 billion in 2024 to \$111.46 billion in 2025.\footnote{\href{https://www.thebusinessresearchcompany.com/report/perishable-prepared-food-global-market-report}{https://www.thebusinessresearchcompany.com/report/perishable-prepared-food-global-market-report/}
, accessed on September 1, 2025.} From a modeling perspective, finite shelf life and lead time uncertainty jointly induce a high-dimensional state that tracks on-hand inventory by remaining lifetime as well as pipeline orders. This structure creates a severe curse of dimensionality, and closed-form optimal policies are generally unavailable.

Human knowledge about policy structure in this theoretically challenging setting is largely derived from simplified models. When demand is allowed to be arbitrary, \citet{chao2015approximation,chao2018approximation} built on the marginal cost accounting framework of \citet{levi2007approximation} to shift cost accounting from period by period to order by order, yielding tractable heuristics with worst-case performance guarantees. Under independent and identically distributed demand, base-stock policies \citep{zhang2020simple,bu2022asymptotic,bu2023managing} and projected inventory level (PIL) policies, which target a desired post-arrival inventory level rather than an ordering-point target \citep{bu2023managing}, are shown to be asymptotically optimal in different regimes. Moreover, \citet{bu2023managing} reported that PIL can outperform base-stock across a wide range of numerical settings. 

Motivated by these two strands of theory, we develop and compare two DNN policies that map observed features and the inventory state to replenishment decisions (Sections~\ref{sec:e2e_bb} and \ref{sec:e2e_pil}). Our first policy, E2E-BB (black-box), is a standard end-to-end network that learns a direct mapping from inputs to order quantities. To enable offline training from historical data, we leverage marginal cost accounting to decompose system cost into decision-level marginal costs, yielding a well-defined training objective. Our second policy, E2E-PIL, further embeds policy structure based on the PIL principle. In this architecture, the inventory state is used only to compute projected on-hand inventory at the replenishment arrival epoch, while the network learns a target level from features without directly using the state. The order quantity is then chosen to move projected inventory toward this learned target, replacing part of the black-box mapping with an interpretable, theory-based mechanism. Thus, while both policies use marginal cost accounting to support end-to-end training, E2E-PIL additionally incorporates inventory-theoretic structure in the network architecture. 

We further show that the objective function induced by E2E-PIL is homogeneous of degree one. This structural property enables us to extend the operational data analytics (ODA) framework \citep{feng2023framework,feng2023transfer,feng2025operational,feng2025contextual} to our dynamic decision-making setting. Leveraging the ODA methodology, we introduce a simple yet effective post-processing step that applies a constant scaling factor to the output of the E2E-PIL policy, where the scaling constant is selected using in-sample data. This ODA-based boosting yields an enhanced policy, which we term E2E-BPIL.

In addition to these E2E approaches, we benchmark against the widely used prediction-then-optimization (PTO) pipeline, which first learns demand and lead time primitives from data and then optimizes decisions in a downstream control model. In particular, we incorporate the proportional-balancing (PB) heuristic \citep{chao2015approximation,chao2018approximation} within this framework, yielding a benchmark we refer to as PTO-PB. We evaluate all methods through extensive numerical studies using a real-world dataset from a beverage company and a suite of synthetic environments that vary demand patterns and lead time characteristics. Across these settings, E2E-PIL consistently attains the lowest cost, with PTO-PB typically ranking second and E2E-BB following. Notably, the two best-performing policies, E2E-PIL and PTO-PB, are precisely those that incorporate inventory-theoretic insights within the E2E and PTO paradigms, respectively, highlighting the practical value of human knowledge for designing high-performing learning-based inventory policies in complex systems.

To understand why E2E-PIL can outperform a more flexible black-box model even though it embeds a policy structure that is not generally optimal in our setting, it is important to recognize the data constraints faced in practice. For many products, replenishment opportunities are infrequent, so even a full year of operations may yield only on the order of a few hundred observations per item. Moreover, as product life cycles shorten and new products are introduced more frequently, older historical data may be less informative for current decisions. In such limited-data environments, model selection is governed by the standard excess risk trade-off in machine learning: richer function classes reduce approximation error but can increase estimation error and harm out-of-sample performance. We formalize this intuition through an excess risk decomposition, showing that embedding an effective heuristic structure reduces policy class complexity and improves learning efficiency while incurring little loss in approximation flexibility. This perspective helps explain the robust empirical advantage of E2E-PIL. More broadly, our findings suggest that deep learning can be made more effective and reliable for operations when it is complemented by domain knowledge, pointing to a productive integration of advanced analytics with established inventory theory.

The remainder of the paper is organized as follows. Section~\ref{sec:review} reviews the related literature. Section~\ref{sec:Model} introduces the problem setup. Section~\ref{sec:method} presents relevant perishable inventory policy structures and develops our proposed end-to-end policies. Section~\ref{sec:num} reports numerical experiments using both real-world and synthetic data. Section~\ref{sec:explanation} provides an explanation based on excess risk decomposition, and Section~\ref{sec:conclusion} concludes.

\section{Literature Review} \label{sec:review}
Our work relates to two streams of literature: perishable inventory management and data-driven algorithms for inventory control.

\textbf{Perishable inventory management}. 
Ordering decisions for perishable products with general lifetimes are challenging because optimal policies are state dependent and quickly become computationally intractable due to the curse of dimensionality \citep{nahmias1975optimal,fries1975optimal}. Consequently, much of the literature develops tractable heuristics \citep[e.g.,][]{nandakumar1993near,sun2014quadratic}. A prominent approach builds on marginal cost accounting: \citet{chao2015approximation} adapted the framework of \citet{levi2007approximation} to propose the proportional-balancing (PB) policy, which trades off marginal holding and outdating costs against marginal backorder costs and admits worst-case performance guarantees for general lifetimes. PB has been extended to settings with setup costs \citep{zhang2016approximation} and with positive lead times and capacity constraints \citep{chao2018approximation}. More recently, \citet{zhangb2023truncated} proposed the truncated balancing policy, which modifies PB by truncating its order quantity using a lower bound to mitigate under-ordering when shortage penalties are high.

A second line of work extends base-stock ideas to perishable systems. Motivated by the strong performance of base-stock policies in nonperishable settings, \citet{zhang2020simple} proposed base-stock heuristic based on fluid and newsvendor approximations and established asymptotic optimality in regimes where the effect of expiration becomes negligible. \citet{bu2022asymptotic} further analyzed asymptotic optimality of base-stock policies across multiple regimes and system variants, including settings with positive lead times. To better account for outdating during the lead time, \citet{bu2023managing} developed the projected inventory level (PIL) policy, which targets a desired inventory level at the replenishment arrival epoch rather than at the ordering epoch.

Our paper contributes to this stream by developing an end-to-end, data-driven approach for perishable inventory systems with random lead times. We leverage marginal cost accounting ideas from \citet{chao2015approximation,chao2018approximation} to construct a decision-level training objective that supports offline policy learning, and we embed the PIL structure of \citet{bu2023managing} into a neural policy class to incorporate interpretable structure beyond a purely black-box model.

\textbf{Data-driven algorithms for inventory management}. 
Most classical inventory policies assume that uncertainty primitives such as demand and lead time distributions are known. In practice, decision makers typically observe only historical data, motivating data-driven approaches. A standard methodology is the two-step prediction-then-optimization (PTO) framework, which first estimates primitives from data and then optimizes decisions in a downstream inventory control model \citep[e.g.,][]{bertsimas2020predictive,cao2024conformal}. We include PTO as a benchmark in our study. Specifically, we first predict conditional means using an off-the-shelf machine learning algorithm, then fit distributions using residual information, and finally plug the fitted distributions into downstream PB decisions following \citet{chao2015approximation,chao2018approximation}.

Beyond PTO, a growing literature integrates prediction and optimization into a unified learning problem. Early work includes operational statistics (OS) for a newsvendor setting without covariates and with unknown demand scale \citep{liyanage2005practical}, which directly integrates data into an explicit decision rule and can outperform parametric approaches and nonparametric sample average approximation. In feature-based newsvendor settings, \citet{ban2019big} studied empirical risk minimization policies with linear decision rules and regularization, while \citet{oroojlooyjadid2020applying} trained nonlinear deep neural network policies by directly minimizing newsvendor costs. \citet{han2023deep} further analyzed excess risk bounds for deep neural networks in the newsvendor problem. Much of this work focuses on single-period settings. For multi-period inventory problems, \citet{qi2023practical} used historical data to construct ex post order quantity labels in a nonperishable system with random lead times and then fit them using a carefully designed neural architecture. Deep reinforcement learning has also been applied to multi-period inventory control problems \citep[e.g.,][]{oroojlooyjadid2022deep,stranieri2025classical,maggiar2025structure}.

Our work differs from prior multi-period learning approaches in two main ways. First, perishability complicates label construction and cost attribution across time, so we build a decision-level training objective using marginal cost accounting rather than relying on ex post labels. Second, rather than relying solely on a black-box network, we embed the PIL principle into the architecture to incorporate policy structure. Finally, \citet{feng2023framework} generalized OS ideas \citep[e.g.,][]{liyanage2005practical,chu2008solving,lu2015technical} into the operational data analytics (ODA) framework, which has been studied primarily in single-period settings \citep{feng2023transfer,chu2025solving,feng2025operational,feng2025contextual}.  We show that our dynamic setting fits this framework and extend the associated boosting idea to further improve performance.

\section{Problem Formulation} \label{sec:Model}

In this section, we describe the problem setting and introduce notations. We consider a finite-horizon perishable inventory system over periods $t=1,...,T$ with stochastic demand and random lead times.  The products have a fixed lifetime of $K$ periods and have zero value after $K$ periods. Unmet demand is backordered and satisfied by future inventory arrivals. Let $D_t$ denote random demand in period $t$. We observe a demand feature vector $\bm{X}^D_t \in \mathbb{R}^{p_1}$ where $p_1$ is the number of demand-related covariates used for forecasting. We model the demand as
\begin{equation*}
    D_t = f^D(\bm{X}^D_t) + \epsilon_t,
\end{equation*}
where $f^D(\cdot)$ is a general function that maps the feature vector to the true mean demand, and $\epsilon_t$ is random noise term with zero mean. 

We assume that the decision maker (DM) periodically reviews the on-hand inventory, but only place replenishment orders at predetermined time points. Specifically, there are $M$ ordering opportunities over the planning horizon, occurring in periods $t_m$ with order quantities $q_m$ for $m = 1, \ldots, M$.  This periodic-ordering structure is consistent with many operational settings (e.g., daily ordering for fresh produce and weekly ordering in retail) and is commonly adopted in the literature \citep[e.g.,][]{qi2023practical}. 

After placing an order at $t_m$, the DM is informed of the delivery lead time before the next ordering point $t_{m+1}$. In perishable supply chains, the need to preserve freshness keeps delivery times within a short operational cycle. Residual lead time uncertainty is largely upstream (availability and processing) and is typically realized soon after order placement, allowing suppliers to provide reliable delivery estimates within the ordering window. Let $L_m$ denote the random lead time associated with the $m$-th order, and let $\bm {X}^L_m \in \mathbb{R}^{p_2}$ be its lead time feature vector, where $p_2$ is the number of covariates. The corresponding arrival period is $v_m = t_m + L_m$.  We model the lead time as
\begin{equation*}
    L_m = f^L(\bm{X}^L_m) + \eta_m,
\end{equation*}
where $f^L(\cdot)$ is a general function that maps the feature vector to the true mean lead time, and $\eta_m$ is a bounded random noise term with zero mean. For any feature vector $\bm{X}^L_m$, we assume that the random lead time is bounded by a constant $\bar{L}$. In addition, we impose the standard no-crossing condition on arrivals, $v_m\leq v_{m+1},\ \forall m=1,...,M-1$ \cite[e.g.,][]{janakiraman2004lost, qi2023practical}. 

Let $\bm z_t \in \mathbb{R}^{K+\bar{L}-1}$ denote the inventory state at the beginning of period $t$, after all pipeline orders have advanced by one period. The state includes both on-hand inventory indexed by remaining lifetime and outstanding orders in the pipeline. Specifically, 
\begin{equation*}
    \bm z_t=(z_{t,1},...,z_{t,K+\bar{L}-1}).
\end{equation*}
For $1 \le i \le K-1$, $z_{t,i}$ is the on-hand quantity at the start of period $t$ with $i$ periods of remaining lifetime.  The component $z_{t,K}$ denotes the net newly available quantity at the beginning of period $t$, defined as the total quantity that arrives at the start of period $t$ minus the backorder carried over from period $t-1$. 
Thus,  $z_{t,K}<0$ indicates that arrivals are insufficient to clear the outstanding backorder. Finally, for $1\leq j\leq \bar L-1$, $z_{t,K+j}$ denotes the quantity of pipeline inventory scheduled to arrive in $j$ periods.

We summarize the system dynamics in our setting through the inventory-state transition from period $t$ to period $t+1$, given in Equation~\eqref{eq:transit}.
\begin{subequations}  \label{eq:transit}
    \begin{align}
        z_{t+1,i} &= \left(z_{t,i+1}-\left(D_{t}-\sum_{j=1}^{i}z_{t,j}\right)^{+}\right)^{+}, &\mathrm{for}\ i=1,\ldots,K-1; \label{eq:transit1} \\
        z_{t+1,i} &=z_{t,i+1}-\left(D_t-\sum_{j=1}^i z_{t,j}\right)^+,&\mathrm{for}\ i=K; \label{eq:transit2} \\
        z_{t+1,i} &=z_{t,i+1},&\mathrm{for}\ i=K+1,\ldots,K+\bar{L}-2; \label{eq:transit3}\\
        z_{t+1,i} &=0,&\mathrm{for}\ i=K+\bar{L}-1; \label{eq:transit4}\\
        z_{t+1,i} &=z_{t+1,i}+q_m,&\mathrm{if}\ t=t_m,\exists m \in \{1,...,M\},i=K+L_m-1. \label{eq:transit5}
    \end{align}
\end{subequations}
In Equation~\eqref{eq:transit1}, on-hand inventory is consumed in first-in-first-out (FIFO) order to fulfill demand (i.e., the oldest items are consumed first), and any remaining units age by one period. Equation~\eqref{eq:transit2} updates the freshest-inventory component $z_{t+1,K}$ for period $t+1$. It equals the scheduled arrival $z_{t,K+1}$ minus the backorder carried over from period $t$, $(D_t-\sum_{j=1}^i z_{t,j})^+$. Thus, $z_{t+1,K}<0$ occurs when the incoming shipment is insufficient to clear the backorder from period $t$.  In Equation~\eqref{eq:transit3} and~\eqref{eq:transit4}, all pipeline inventory advances one period closer to delivery. Finally, in Equation~\eqref{eq:transit5}, the new replenishment is added to the appropriate pipeline position: it is inserted into the pipeline when $L_m>1$ and contributes to next period’s arrival when $L_m=1$. Note that when $L_m=1$, $z_{t+1, K}$ includes both the pre-existing one-period-ahead pipeline quantity $z_{t,K+1}$ and the newly placed order $q_m$.

During each period, the system may incur three types of costs: 1) \emph{Outdating Cost:} If the quantity of items with one period of remaining lifetime exceeds the demand (i.e., $z_{t,1} > D_t$), the excess inventory will expire and incur an outdating cost of $\theta(z_{t,1} - D_t)^+$ where $\theta$ denotes per-unit outdating cost. 2)  \emph{Holding Cost:} If the total on-hand inventory exceeds the demand (i.e., $\sum_{i=1}^K z_{t,i} > D_t$), the surplus will be subject to a holding cost of $h\left(\sum_{i=1}^K z_{t,i} - D_t\right)^+$ where $h$ denotes per-unit holding cost. 3) \emph{Backorder Cost:} If the total on-hand inventory is less than the demand (i.e., $\sum_{i=1}^K z_{t,i} \le D_t$), the unmet demand will be backordered and incur a backorder cost of $b\left(D_t - \sum_{i=1}^K z_{t,i}\right)^+$ where $b$ denotes per-unit backorder cost.  At the end of the horizon $T$, any
remaining inventory has zero salvage value. For simplicity, we do not include ordering costs because a standard cost transformation can accommodate a positive per-unit ordering cost \citep{chao2015approximation, chao2018approximation}.

The DM’s objective is to choose an ordering policy $\pi$  specifying the order quantity $q_m$ at each ordering epoch $t_m$ to minimize the expected total cost over the horizon. Under policy $\pi$, the expected total cost is
\begin{equation} \label{eq:objective}
     \mathbb{E} \left[\sum_{t=1}^{T}C^\pi_t\right] = \mathbb{E} \left [\sum_{t=1}^{T}\left (h\biggl(\sum_{i=1}^Kz^\pi_{t,i}-D_t\biggr)^++b\biggl(D_t-\sum_{i=1}^Kz^\pi_{t,i}\biggr)^++\theta\biggl(z^\pi_{t,1}-D_t\biggr)^+\right )\right ],
\end{equation}
where $C^\pi_t$ is the realized cost in period $t$ under policy $\pi$, and $z^\pi_{t,i}$ is the $i$-th component of $\bm z_t$ induced by policy $\pi$. For simplicity, we assume the initial state $\bm z_1\equiv\bm 0$; all results extend to an arbitrary initial inventory state.

The optimal policy for minimizing the objective in Equation~\eqref{eq:objective} is computationally intractable due to the curse of dimensionality, even when the demand and lead time distributions are fully specified. In practice, this challenge is magnified because the functional forms of $f^D(\cdot)$ and $f^L(\cdot)$ and the stochastic processes of $\epsilon_t$ and $\eta_m$ are not known a priori. Accordingly, our goal is to learn an ordering policy from historical data that achieves strong out-of-sample performance.

To achieve this, we construct training samples from historical observations at ordering epochs. For each historical ordering point $t_m$ where $m<1$ to represent the history, we extract fixed-length windows of past demand and lead time realizations, together with their associated features, and concatenate them into an input vector $\bm{x}_{t_m}\in\mathbb{R}^p$, where $p$ denotes the input dimension. The corresponding labels are constructed from future windows. The future demand target is $\bm{d}_{t_m}=\{d_t\}_{t=t_m}^{t_m+K+\bar{L}-1}$ and the future lead time target is $\bm{l}_m=\{l_m,l_{m+1}\}$. The future demand window extends to period $t_m+K+\bar{L}-1$ which is the latest period at which the order placed at $t_m$ may affect demand fulfillment. Therefore, we can use the future demand target $\bm{d}_{t_m}$ to measure the cost induced by the order quantity $q_m$  in period $t_m$, and use this cost signal to guide the training of our model. We include two future lead times to capture the interval between current and subsequent arrivals which determines the demand the current order must cover. Appendix~\ref{ecsec:data_arrange} provides details on this data organization. In Section~\ref{subsubsec:loss}, we introduce  how to construct suitable inventory state samples for training, noting that the inventory state depends on the ordering policy.

\section{The End-to-End Policy Learning} \label{sec:method}

Our objective is to learn an end-to-end (E2E) ordering policy that maps the information available at ordering epochs, 
including covariates and the current inventory state, to an order quantity. 
Conceptually, an E2E policy approximates the optimal dynamic programming mapping directly, rather than following the traditional two step approach that first estimates demand and lead time distributions and then solves an optimization problem. However, training E2E policies in perishable systems faces two  difficulties.

First, a single order has intertemporal cost consequences in a dynamic perishable inventory system. An order placed at $t_m$ may arrive several periods later and may partially satisfy demand over multiple periods before expiring. As a result, the cost impact attributable to that order propagates through holding, outdating, and backorder costs over an extended future window. This temporal coupling motivates dynamic programming, but it complicates learning because the training signal for the decision at $t_m$ is not directly observable from outcomes within period $t_m$. One potential remedy is reinforcement learning, which can incorporate the system dynamics explicitly. However, offline reinforcement learning with high-dimensional states and delayed costs can be unstable and sample intensive when only historical trajectories are available. In nonperishable settings, for example, \citet{qi2023practical} sought to avoid reinforcement learning by recovering ex post optimal order quantities from realized future demands and lead times and then fitting these labels using a distance based loss within a deep learning framework. In our perishable setting, however, inventory aging and the interaction between backorder clearing and pipeline arrivals make such ex post labeling substantially more complex and typically not computationally efficient.

To address the intertemporal cost impact of a single order, we rely on the marginal cost accounting scheme, introduced by \citet{levi2007approximation} and adapted to perishable systems by  \citet{chao2015approximation,chao2018approximation}. The key idea is to reexpress the total cost by charging each order the future costs that are “caused” by that order, rather than charging costs period by period. In particular, once an order quantity is chosen, the outdating, holding, and backorder costs attributable to that order are determined by the current state and the realized demand path, and they do not depend on subsequent ordering decisions. This yields a per-order, decision-aligned cost that serves as a natural loss function for training an E2E policy within a deep learning framework.

Second, the state-to-action mapping is high-dimensional. A generic deep learning architecture permits a highly flexible dependence of the order quantity on the inventory state. As a result, even though the system dynamics impose strong structure, such as FIFO issuance and deterministic aging, a purely black-box model does not exploit this structure in a systematic way and must learn it from data. Without structural guidance, model capacity is devoted to rediscovering relationships that are well understood in inventory theory, which can increase sample requirements and reduce out-of-sample performance.

To address this issue, we leverage the projected inventory level (PIL) policy proposed by \citet{bu2023managing}, which provides an effective and economically interpretable ordering structure for perishable systems. Rather than targeting a base stock level at the ordering epoch, PIL raises the expected inventory level at the replenishment arrival epoch to a target level. Under a stationary setting, \citet{bu2023managing} established several asymptotic optimality results for the PIL policy and document strong empirical performance across a range of settings. Embedding the PIL structure into the network architecture reduces the effective complexity of the state-to-action mapping, thereby lowering the learning burden and improving robustness when data are limited or the state dimension is large.

In the remainder of this section, we first introduce the marginal cost accounting scheme and use it to construct the loss function for policy training in Section~\ref{sec:mcas_loss}. Based on this loss function, we then propose the end-to-end black-box (E2E-BB) policy in Section~\ref{sec:e2e_bb} which learns the full policy mapping using a DNN. Finally, in Section~\ref{sec:e2e_pil}, we review the PIL policy and describe how we embed its structure into another network architecture.

\subsection{Marginal Cost Accounting Scheme and the Loss Function} \label{sec:mcas_loss}

We first review the marginal cost accounting scheme for perishable systems, following \citet{chao2015approximation,chao2018approximation}, in Section~\ref{subsubsec:mcas}, and then develop the loss function used for policy learning in Section~\ref{subsubsec:loss}.

\subsubsection{Marginal Cost Accounting Scheme} \label{subsubsec:mcas}

The marginal cost accounting scheme reformulates the total cost over the planning horizon by moving from a time based accumulation of costs to a decomposition into marginal costs associated with individual orders. We adapt the scheme to fit our specific setting. For detailed derivations and further explanation, we refer readers to \citet{chao2015approximation, chao2018approximation}.

We begin by introducing two auxiliary variables that capture the inventory dynamics as follows:

\textbf{\textit{Auxiliary Variables.}}
For $t_m\leq s\leq v_m+K-2$, we define $B_{[t_m,s]}(\bm{z}_{t_m})$ as the quantity of items which outdate between periods $[t_m,s]$ for  the given inventory vector $\bm{z}_{t_m}$ in period $t_m$. It is computed recursively as
\begin{equation}\label{eq:outdata_quantity}
    B_{[t_m,s]}(\bm{z}_{t_m})=\max \left \{\sum_{i=1}^{s-t_m+1}z_{t_m,i}
-D_{[t_m,s]}, B_{[t_m,s-1]}(\bm{z}_{t_m})\right \},
\end{equation}
where $D_{[t_m,s]}=\sum_{t=t_m}^sD_t$, and we let $B_{[t_m,t_m-1]}(\bm{z}_{t_m})\equiv0$ for completeness. Next, for $t_m\leq s\leq v_m+K-1$, we define $\tilde{D}_s(\bm{z}_{t_m})$ as the unmet demand after only consuming the inventory available in $\bm{z}_{t_m}$ and without using $q_m$ and subsequent orderings. It is given by
\begin{equation} \label{eq:d_tilde}
    \tilde{D}_s(\bm{z}_{t_m}):=D_{[t_m,s]}-\left [
\sum_{i=1}^{K+\bar{L}-1}z_{t_m,i}-B_{[t_m,s-1]}(\bm{z}_{t_m})\right ],
\end{equation}
where $\tilde{D}_s(\bm{z}_{t_m})<0$ indicates that inventory from $\bm{z}_{t_m}$ remains available at period $s$. Note that because we assume that there is no crossing-over of orders, then $\sum_{i=1}^{K+\bar{L}-1}z_{t_m,i}=\sum_{i=1}^{K+L_m-1}z_{t_m,i}$ for any realized $L_m$.

Using the auxiliary variables $B_{[t_m,s]}(\bm{z}_{t_m})$ and $\tilde{D}_s(\bm{z}_{t_m})$, we derive the marginal holding cost, marginal outdating cost, and marginal backorder cost for each order quantity, conditioned on the current inventory state.

\textbf{\textit{Marginal Holding Costs.}}
The marginal holding cost of the $m$-th order $q_m$, given the current inventory state $\bm{z}_{t_m}$, can be incurred throughout the product’s life cycle from its arrival period $v_m$ to the last period in which units from that order can still be held and used, which is $(v_m + K - 1) \land T$, where $t\land t' = \min(t,t')$. Inventory from the new order is only consumed when the remaining inventory from previous orders is depleted, i.e., when $\tilde{D}_s(\bm{z}_{t_m}) > 0$. The marginal holding cost is thus given by 
\begin{equation*} H(q_{m}|\bm{z}_{t_m}):= h\sum_{s=v_m}^{(v_{m}+K-1)\land T}\left (q_m- \left (\tilde{D}_s(\bm{z}_{t_m})\right )^+\right )^+. \end{equation*}

\textbf{\textit{Marginal Outdating Costs.}}
The marginal outdating cost of the $m$-th order $q_m$, given the current inventory state $\bm{z}_{t_m}$, arises only if some units from this order remain unconsumed at the end of its shelf life, i.e., in period $v_m + K - 1$. When $v_m + K - 1 \leq T$,  the marginal outdating cost is 
\begin{equation*} \Theta(q_{m}|\bm{z}_{t_m}):=\bm{1}\{v_m+K-1\leq T\}\cdot \theta\left(q_{m}-\left (\tilde{D}_{v_m+K-1}(\bm{z}_{t_m})\right)^+\right)^+. \end{equation*}

\textbf{\textit{Marginal Backorder Costs.}} The marginal backorder cost of the $m$-th order $q_m$, given the current inventory state $\bm{z}_{t_m}$, arises when the order is active (i.e., after its arrival and before either its expiration or the arrival of the subsequent order) but is insufficient to fully satisfy the demand for which it is responsible. Formally, the marginal backorder cost is given by
\begin{equation*} \Pi(q_m|\bm{z}_{t_m}):=b\sum_{s=v_m}^{(v_{m+1}-1)\land (v_m+K-1)\land T} \left(\tilde{D}_s(\bm{z}_{t_m})-q_m\right)^+.
\end{equation*}

Note that all auxiliary variables and marginal costs implicitly depend on future demands and lead times and are therefore random variables. According to~\citet{chao2015approximation,chao2018approximation}, the marginal cost accounting scheme is equivalent to the original period by period accounting in the sense that it yields the same realized total cost. In particular, the total cost in Equation~\eqref{eq:objective} can be rearranged as
\begin{equation} \label{eq:mcas}
    \sum_{t=1}^{T}C^{\pi}_t=\sum_{m=1}^M \left [H(q^\pi_{m}|\bm{z}^\pi_{t_m})+ \Theta(q^\pi_{m}|\bm{z}^\pi_{t_m})+\Pi(q^\pi_{m}|\bm{z}^\pi_{t_m}) + b\cdot\sum_{s=(v_{m+1})\land (v_m+K)\land T}^{(v_{m+1}-1)\land T} D_s\right]+\sum_{t=1}^{L_1}C^{\pi}_t.
\end{equation}
Equation~\eqref{eq:mcas} includes two additional terms beyond the marginal costs. The term inside the square brackets captures the residual backorder cost incurred after the $m$-th order has expired (i.e., after period $v_m + K$) and before the next order arrives (i.e., before $v_{m+1}$). The final term $\sum_{t=1}^{L_1}C^{\pi}_t$ represents costs incurred during the initial lead time, before any order can arrive. Both terms are independent of the ordering decisions and thus cannot be influenced by the policy. Accordingly, we omit them in the subsequent analysis.

\subsubsection{Marginal Cost as the Loss Function} \label{subsubsec:loss}

The marginal cost accounting scheme isolates the marginal impact of each decision conditional on the current state, and thus yields a per-decision loss function. Specifically, for a candidate order quantity $\hat{q}_m$ at state $\bm z_{t_m}$, we define the loss as the realized total marginal cost associated with this order:
\begin{equation} \label{eq:loss}
\mathcal{L}\left(\hat{q}_m|\bm{z}_{t_m}\right):=H\left(\hat{q}_{m}|\bm{z}_{t_m}\right)+\Theta\left(\hat{q}_{m}|\bm{z}_{t_m}\right)+\Pi\left(\hat{q}_m|\bm{z}_{t_m}\right).
\end{equation}
The loss function defined in Equation \eqref{eq:loss} is computationally tractable, convex, and almost everywhere differentiable, enabling efficient training of the E2E model. The E2E policy aims to approximate the optimal mapping from the entire inventory state space to ordering decisions. A practical challenge, however, is that historical datasets often do not contain inventory state records. Even when state records are available, they typically correspond to a single realized trajectory generated by a particular policy, which provides limited coverage of the state space and hinders generalization beyond the observed path.

To address this challenge, we decouple policy learning from the single inventory trajectory observed in operations. Instead of using only the states generated by the historical policy, we pre-sample a representative set of inventory states prior to training. Specifically, we simulate the system using the in-sample demand and lead time data under a simple policy, such as the benchmark described in Section~\ref{sec:num}, and collect the resulting inventory vectors ${\bm z_{t_m}}$. This procedure produces a realistic distribution of states that provides broader coverage of the relevant region of the state space. The approach is inspired by constraint sampling methods in approximate dynamic programming \citep{de2004constraint,farias2007approximate}, which approximate large state spaces using a carefully selected subset of states. Pairing these sampled states with the corresponding realizations of future demands and lead times allows us to compute $\mathcal{L}(\hat{q}_m | \bm z_{t_m})$, enabling efficient training of the E2E policy model.

\subsection{The E2E Black-Box Policy without Enforcing Inventory Policy Structure} \label{sec:e2e_bb}

We first introduce a fully black-box end-to-end order quantity policy, denoted as E2E-BB, where BB stands for black-box and no inventory policy structure is imposed. To mitigate overfitting and encourage the network to learn informative representations, we augment the training objective with auxiliary prediction losses for demand and lead time. The resulting loss function of the E2E-BB policy, implemented via a DNN, is given by:
\begin{align} \label{eq:loss_e2e_bb}
\mathcal{L}^{\text{E2E-BB}}\!\left(\hat{q}_m\!\left(\bm x_{t_m},\bm z_{t_m}\right),\hat{\bm d}\!\left(\bm x_{t_m}\right),\hat{\bm l}\!\left(\bm x_{t_m}\right)\middle|\bm z_{t_m}\right) 
= &\mathcal{L}\Bigl(\hat{q}_m(\bm x_{t_m},\bm z_{t_m})\Big|\bm z_{t_m}\Bigr) 
+ \lambda^{D}\ell^D\!\left(\hat{\bm d}_{t_m}\!\left(\bm x_{t_m}\right),\bm d_{t_m}\right) \nonumber \\
&+ \lambda^{L}\ell^L\!\left(\hat{\bm l}_{m}\!\left(\bm x_{t_m}\right),\bm l_{m}\right),
\end{align}
where $\ell^D$ and $\ell^L$ are distance-based loss functions (e.g., mean squared error), and $\lambda^{D}$ and $\lambda^{L}$ are hyperparameters that control the regularization strength.

\begin{figure}[htbp]
     \FIGURE
    {\includegraphics[width=0.75\linewidth]{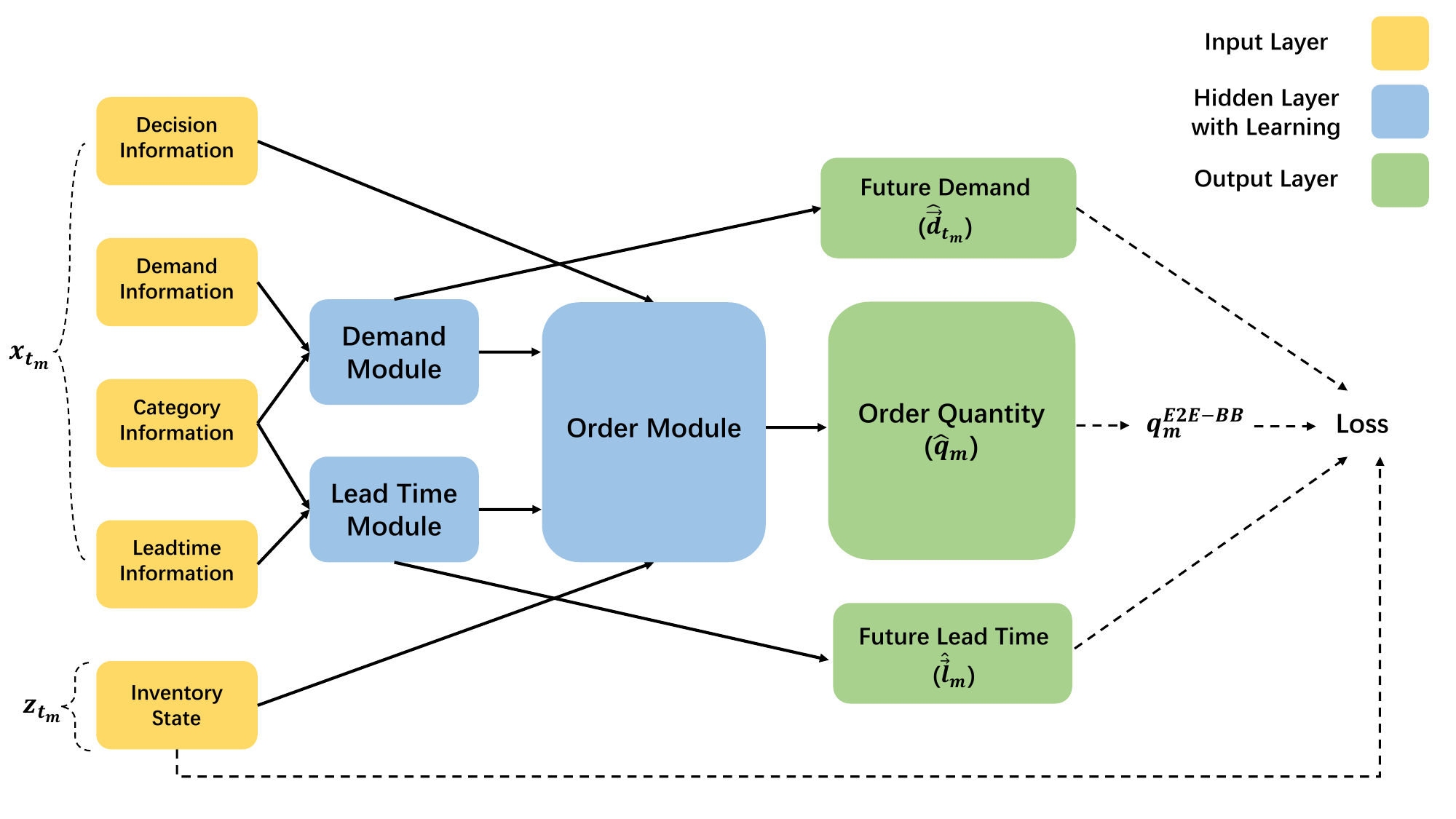}} 
{The deep neural network architecture of the E2E-BB policy. \label{fig:e2e-bb}}
{}
\end{figure}
We design a DNN architecture for the E2E-BB policy, as illustrated in Figure~\ref{fig:e2e-bb}. The primary objective of the model is to approximate the optimal order quantity. To facilitate the training process and provide regularization via auxiliary loss terms, the network additionally outputs predictions of future demands and lead times. The model takes two types of inputs: the feature vector $\bm x_{t_m}$ and the current inventory state $\bm z_{t_m}$. The feature vector $\bm x_{t_m}$ includes decision-relevant information such as simple benchmark guidance (e.g., the empirical quantile of historical demand multiplied by average historical lead time \citep{qi2023practical}) and weekday indicators, demand information, product category information, and lead time information.

The network processes $\bm x_{t_m}$ through two specialized components: the Demand Module and the Lead Time Module, which learn representations tailored to their respective prediction tasks. Their representations are then combined with $\bm z_{t_m}$ in an Order Module to form a joint representation for decision making. The network produces three outputs: predicted future demands ($\hat{\bm d}_{t_m}$), predicted future lead times ($\hat{\bm l}_m$), and the key decision variable, the recommended order quantity ($q_m^{\text{E2E-BB}}$) which defines the E2E-BB replenishment policy. We denote the resulting policy by $\pi^{\text{E2E-BB}}(\bm x_{t_m},\bm z_{t_m})$, thereafter.

\subsection{The E2E Projected Inventory Level Policy} \label{sec:e2e_pil}

In Section~\ref{sec:e2e_bb}, we introduced a baseline E2E-BB policy that maps the observable feature vector $\bm{x}_{t_m}$ and the inventory state vector $\bm{z}_{t_m}$ to the order quantity $q_m$ using a DNN. This approach treats the mapping as a generic black-box function, without explicitly incorporating structural insights from classical inventory theory. However, in many canonical settings, the relationship between $\bm{z}_{t_m}$ and $q_m$ is analytically tractable and known to be (asymptotically) optimal. For instance, under a nonperishable inventory system with constant lead times, the base-stock policy is provably optimal, enforcing that the sum of the inventory position and the order quantity remains at a fixed target level.

For perishable products with fixed lifetimes and constant lead times, \citet{bu2023managing} studied a variant of the base-stock policy, termed the projected inventory level (PIL) policy, which raises the expected inventory position at replenishment \emph{arrival} epoch to a target level. They establish the asymptotic optimality results and document strong empirical performance across a range of scenarios. Motivated by this structure, we design a theory-guided E2E policy that embeds the PIL structure within the end-to-end learning framework.

\subsubsection{The Projected Inventory Level Policy} \label{sec:pil}

We first review the PIL policy \citep{bu2023managing} and then discuss how to embed it into the network in Section~\ref{subsubsec:embed_pil}. For the $m$-th order, the DM aims to raise the expected inventory position upon the order arrival (i.e., at the beginning of period $v_m$) to a target level $S_m$. To achieve this, the DM must place an order with quantity $q_m$ calculated as follows:
\begin{equation*}
q_m=\left(S_m-\mathbb E\left[\sum_{i=1}^{K}\tilde{z}_{v_m,i}\middle| \bm z_{t_m}\right]\right)^+,
\end{equation*}
where $\tilde{\bm z}_{v_m}$ denotes the \emph{counterfactual} on-hand inventory level at period $v_m$, obtained by evolving the state transitions in Equation~\eqref{eq:transit} under future demand and lead time realizations, under a counterfactual scenario where the current order is ignored (i.e, $q_m=0$). Expanding the transitions and utilizing the auxiliary variable defined in Equation~\eqref{eq:outdata_quantity}, the resulting counterfactual on-hand inventory at the beginning of period $v_m$ can be written as:
\begin{equation*}
\sum_{i=1}^{K}\tilde{z}_{v_m,i}=\sum_{i=1}^{K+\bar L-1}z_{t_m,i}-D_{[t_m,v_m-1]}-B_{[t_m,v_m-1]}(\bm z_{t_m}).
\end{equation*}
This equation shows that the projected on-hand inventory at period $v_m$ is the total inventory in the current state $\bm z_{t_m}$ minus the cumulative demand satisfied $D_{[t_m,v_m-1]}$ and the inventory $B_{[t_m,v_m-1]}(\bm z_{t_m})$ that becomes outdated. For notational simplicity and consistency with the term ``projected inventory level", we 
define the projected on-hand inventory (POI) as $z_{v_m}^{\text{POI}}=\sum_{i=1}^{K}\tilde{z}_{v_m,i}$. 

Note that when $K \rightarrow \infty$ (i.e., the nonperishable case), we have $B_{[t_m,v_m-1]}(\bm z_{t_m})\equiv 0$, and the PIL policy reduces to the classical base-stock policy, with the corresponding base-stock level $S_m+\mathbb E[D_{[t_m,v_m-1]}]$. In the perishable case, the PIL policy explicitly accounts for outdating during the lead time and thereby exercises more direct control over the inventory states at the arrival points, achieving superior performance compared to the base-stock policy in most scenarios, as demonstrated by \citet{bu2023managing}.

\subsubsection{Embed the Projected Inventory Level Structure into the Network} \label{subsubsec:embed_pil}

\begin{figure}[htbp]
     \FIGURE
    {\includegraphics[width=0.8\linewidth]{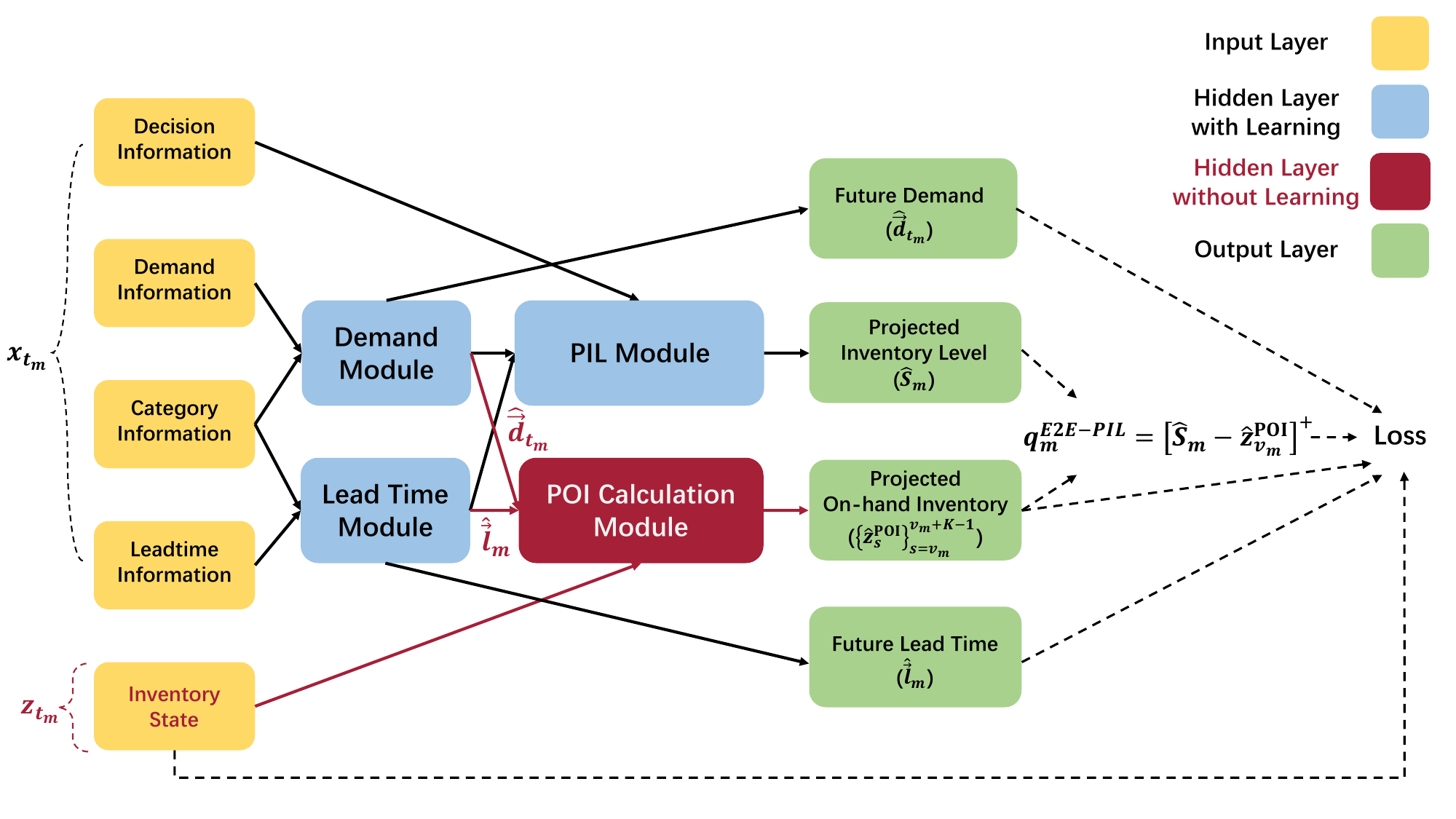}} 
{The deep neural network architecture of the E2E-PIL policy. \label{fig:e2e-pil}}
{}
\end{figure}

Building upon the PIL policy reviewed above, we now introduce the end-to-end projected inventory level (E2E-PIL) policy, which integrates the analytical structure of the PIL policy into a neural network. As illustrated in Figure~\ref{fig:e2e-pil}, the key distinction from the E2E-BB architecture (Figure~\ref{fig:e2e-bb}) lies in how the inventory state $\bm{z}_{t_m}$ is handled. Instead of being processed by a learnable module, $\bm z_{t_m}$ is mapped through a deterministic, domain-knowledge-based calculation block, which reduces the learning burden on the neural network.

Under the E2E-PIL framework, the network recommends the target PIL $\hat{S}_m$ for the order arrival epoch and estimates the POI $\hat{z}^{\text{POI}}_{v_m}$ through an embedded calculation module. The resulting order quantity is then
\begin{equation*}
q_m^{\text{E2E-PIL}} = [\hat{S}_m - \hat{z}^{\text{POI}}_{v_m}]^+.
\end{equation*}
Noting that, beyond recommending the projected inventory level $\hat{S}_m$, computing $\hat{z}^{\text{POI}}_{v_m}$ also requires knowledge of future information, such as the demand sequence from $t_m$ to $v_m-1$ and the lead time. To accelerate this computation, \citet{bu2023managing} proposed approximating the expected POI using the certainty-equivalence path, which assumes that all random lead-time demands take their expected values. Following this approach, at time $t_m$, we calculate $\hat{z}^{\text{POI}}_{v_m}$ based on the predicted demand sequence $\hat{\bm d}_{t_m}$ and the predicted lead time $\hat{\bm l}_m$, where these predictions serve as estimations of the true expectations of the demands and lead time.

Building on Equation~\eqref{eq:loss_e2e_bb}, we augment the training objective by adding a prediction loss of the POI, on top of the existing loss components. The resulting loss function for the E2E-PIL policy is:
\begin{align}
\mathcal{L}^{\text{E2E-PIL}} \Bigl(
    \hat{S}_m\!\left(\bm x_{t_m}\right),
    \hat{\bm d}\!\left(\bm x_{t_m}\right),
    \hat{\bm l}\!\left(\bm x_{t_m}\right),
    \hat{\bm z}^{\text{POI}}\!\left(\bm x_{t_m}, \bm z_{t_m}\right)
\Bigr) \nonumber 
&= 
\mathcal{L}\!\left(
    \Bigl[\hat{S}_m\!\left(\bm x_{t_m}\right)
    - \hat{z}^{\text{POI}}_{v_m}\!\left(\bm x_{t_m}, \bm z_{t_m}\right)\Bigr]^+
    \,\middle|\, \bm z_{t_m}
\right) \nonumber \\
&\quad + \lambda^D \, \ell^D\!\left(\hat{\bm d}_{t_m}\!\left(\bm x_{t_m}\right), \bm d_{t_m}\right)
      + \lambda^L \, \ell^L\!\left(\hat{\bm l}_{m}\!\left(\bm x_{t_m}\right), \bm l_{m}\right) \nonumber \\
&\quad + \lambda^{\text{POI}}_1 \,
      \ell^{\text{POI}}_1\!\Bigl(
        \hat{z}^{\text{POI}}_{v_m}\!\left(\bm x_{t_m}, \bm z_{t_m}\right),
        z^{\text{POI}}_{v_m}
      \Bigr) \nonumber \\
&\quad + \lambda^{\text{POI}}_2
      \sum_{i=1}^{K-1}
      \ell^{\text{POI}}_2\!\Bigl(
        \hat{z}^{\text{POI}}_{v_m+i}\!\left(\bm x_{t_m}, \bm z_{t_m}\right),
        z^{\text{POI}}_{v_m+i}
      \Bigr).
\end{align}
Here, $\lambda^{\text{POI}}_1$ and $\lambda^{\text{POI}}_2$ are hyperparameters. The term $\ell^{\text{POI}}_1(\hat{z}^{\text{POI}}_{v_m}, z^{\text{POI}}_{v_m})$ captures the prediction loss of the POI at time $v_m$, while the summation $\sum_{i=1}^{K-1} \ell^{\text{POI}}_2(\hat{z}^{\text{POI}}_{v_m + i}, z^{\text{POI}}_{v_m + i})$ captures the prediction loss over the subsequent time window during which the current order $q_m$ may affect system dynamics. To ensure that these losses reflect forecasting/simulation errors rather than differences induced by ordering decisions, all POI targets and predictions (i.e., $\{ z^{\text{POI}}_{v_m + i} \}_{i=0}^{K-1}$ and $\{ \hat{z}^{\text{POI}}_{v_m + i} \}_{i=0}^{K-1}$) are all computed under a counterfactual “no-order” rollout that ignores the current and all subsequent orders (i.e., setting $q_{m'}=0$ for all $m'\ge m$). Additional implementation details for computing POI values are provided in Appendix~\ref{ecsec:POI_cal}.

We now formally introduce the tailored neural network architecture for the E2E-PIL policy. As illustrated in Figure~\ref{fig:e2e-pil}, the demand and lead time prediction modules retain the same structure as in the E2E-BB policy, producing sequences of predicted demands ($\hat{\bm d}_{t_m}$) and lead times ($\hat{\bm l}_{m}$), respectively. However, unlike E2E-BB, the PIL module in E2E-PIL outputs the predicted PIL ($\hat{S}_m$) instead of directly generating the order quantity. The key innovation in this architecture is an embedded POI calculation module, which deterministically simulates the system dynamics to produce the predicted POI trajectory $\{ \hat{z}^{\text{POI}}_s \}_{s=v_m}^{v_m + K - 1}$ based on $\hat{\bm d}_{t_m}$, $\hat{\bm l}_{m}$, and the current inventory state $\bm z_{t_m}$. The final order decision is then given by $q_m^{\text{E2E-PIL}} = [\hat{S}_m - \hat{z}_{v_m}^{\text{POI}}]^+$. We denote this E2E-PIL policy as $\pi^{\text{E2E-PIL}}(\bm x_{t_m},\bm z_{t_m})$.

\subsubsection{The Boosted E2E Projected Inventory Level Policy} \label{sec:oda}

In this section, we propose a boosted variant of the E2E-PIL policy by leveraging the idea of operational data analytics (ODA) \citep[e.g.,][]{feng2023framework,feng2023transfer,feng2025operational,feng2025contextual,chu2025solving}. The key ODA insight is that, when the underlying objective is (positively) homogeneous, one can improve a candidate data-driven policy through a simple multiplicative calibration. Specifically, if scaling both the exogenous randomness (e.g., demand) and the decision by a factor $\gamma>0$ scales the resulting cost by the same factor, then the relative structure of a good policy is preserved under scaling, and the residual error of a learned policy is often primarily a mismatch in magnitude rather than in shape. This motivates augmenting the E2E-PIL policy with a one-parameter ``boosting'' layer that outputs $\gamma\cdot \pi^{\text{E2E-PIL}}(\cdot)$, where $\gamma$ serves as a parsimonious knob to correct systematic over- or under-ordering without relearning the full policy.

We first verify that this homogeneity property holds in our dynamic inventory setting. Let $C(\{D_t\}_{t=1}^T,\{q^\pi_m\}_{m=1}^M)$ denotes the total cost under policy $\pi$, written explicitly as a function of the realized demand sequence and the induced decision sequence. 
\begin{proposition} \label{prop_oda}
    The total cost $C(\{D_t\}_{t=1}^T,\{q^\pi_m\}_{m=1}^M)$ is homogeneous of degree $1$. That is, for any demand sequence $\{D_t\}_{t=1}^T$, any policy $\pi$ with decision sequence $\{q_m^\pi\}_{m=1}^M$, and for any $\gamma \in \mathbb R_+$, we have
    \begin{equation*}
        C(\{\gamma D_t\}_{t=1}^T,\{\gamma q^\pi_m\}_{m=1}^M)=
        \gamma C(\{D_t\}_{t=1}^T,\{q^\pi_m\}_{m=1}^M).
    \end{equation*}
\end{proposition}

Proposition~\ref{prop_oda} follows because both the one-step transition mapping and the one-period cost are positively homogeneous, and backward induction then transfers this property to the multi-period objective. Leveraging this insight, we enrich the expressive power of our learned policy by introducing a boosted class of policies, which scales the output of the trained E2E-PIL policy by a positive parameter $\gamma$:
\begin{equation*}
    \mathcal{F}_{\text{B:PIL}} = \left\{ f: \mathbb{R}^p \times \mathbb{R}^{K + \bar{L} - 1} \times \mathbb{R}_+ \rightarrow \mathbb{R}\ \middle | \ f(\bm{x}, \bm{z}, \gamma) = \gamma \cdot \pi^{\text{E2E-PIL}}(\bm{x}, \bm{z}),\ \gamma \in \mathbb{R}_+ \right\},
\end{equation*}
and search for the optimal scaling parameter $\gamma^*$ within this class.

Although we do not derive a theoretical value for $\gamma^*$ due to the unknown distribution, we tune it empirically by simulating in-sample dynamics and selecting the value that minimizes the in-sample cost. We refer to this enhanced version as the E2E-BPIL (Boosted PIL) policy. 

\section{Numerical Study} \label{sec:num}

In this section, we present numerical experiments to evaluate the performance of the proposed policies. In addition to E2E policies introduced in Section~\ref{sec:method}, we incorporate the PB heuristic \citep{chao2015approximation,chao2018approximation} into the PTO framework as a benchmark, referred to as the PTO-PB policy. Specifically, we first predict the conditional means using an off-the-shelf machine learning algorithm, then fit multivariate normal distributions for demand and (rounded) lead time based on the residuals, and finally plug the fitted distributions into the downstream PB heuristics. Additional implementation details are provided in Appendix~\ref{ecsec:benchmark_pb}.

In Section~\ref{num:real}, we introduce the real-world dataset and evaluate all the aforementioned policies to assess their practical effectiveness. In Section~\ref{num:syn}, to further demonstrate robustness, we conduct extensive comparisons using synthetic data under diverse configurations. All experiments are conducted on a personal computer equipped with a 12th Generation Intel(R) Core(TM) i5-12400F CPU and an NVIDIA GeForce RTX 4060 Ti GPU. The implementation uses Python 3.12.3 and PyTorch 2.3.1. The source code is publicly available on GitHub\footnote{Refer to repository: \href{https://github.com/Xuansjtuantai/e2e_perishable_inventory.git}{https://github.com/Xuansjtuantai/e2e\_perishable\_inventory.git} }.

\subsection{Primary Investigation based on a Real-World Dataset} \label{num:real}

We utilize a dataset on demand and lead time, obtained from an open data challenge\footnote{\href{https://www.coap.online/competitions/1}{https://www.coap.online/competitions/1}} . The dataset is provided by a major beverage supplier that operates its own distribution centers (DCs) to hold inventory and fulfill demand of various products across different regions through its managed supply chain network. We focus on perishable products with short shelf lives, such as fresh beer and fresh milk.

\subsubsection{Data Description and Performance Metric}  \label{sec:datadescript}

In the company's supply chain network, DCs hold inventory and fulfill delivery commitments to downstream customers, including e‑commerce platforms, supermarket retailers, and restaurants. On a predetermined schedule, each DC periodically reviews its stock levels and replenish its stock keeping units (SKUs) from upstream factories. Because production may be delayed, replenishment lead times are uncertain. Demand is fulfilled from on-hand inventory in accordance with service agreements. If a stockout occurs, the DC incurs a revenue loss and backorders any unfilled items, delivering them as soon as replenishment arrives.

According to the data, the company operates $J=18$ DC and supplies $I=77$ different SKUs. The same SKU can exhibit different demand and lead time distributions across  DCs due to regional variations. To account for this heterogeneity, we treat each SKU–DC pair as a distinct product. Throughout the paper, we use the terms “product” and “SKU–DC pair”, interchangeably. After data cleaning, our dataset contains daily sales and lead time records for 853 SKU–DC pairs from January 2019 to March 2019, yielding a planning horizon of $T = 90$ days. We use the first $T^{\text{in}} = 60$ days for in-sample training and the remaining $T^{\text{out}} = 30$ days for out-of-sample evaluation. The detailed descriptive analysis of the dataset can be found in Appendix \ref{ecsec:descriptive}.

In our numerical setup, we assume that orders are placed at fixed intervals, denoted by $R$. To fully utilize the realized demand and lead time data, we generate $R$ simulation paths by shifting the ordering epochs so that every data point can serve as a potential ordering point. Specifically, for each $r \in \{1, \dots, R\}$, the ordering periods in the $r$-th simulation path are given by $\{r, r+R, r+2R, \dots\}$.

The original dataset does not include cost parameters. Therefore, we adopt the commonly used values from \citet{chao2018approximation}, setting unit holding cost $h=1$, unit backorder cost $b=10$, and unit outdating cost $\theta=10$. Since the no-crossing-over assumption cannot be guaranteed in the data, we set the ordering interval to $R=4$ to achieve no crossing-over with high probability. Even if crossovers occur with low probability, our method remains applicable, though it may incur corresponding losses. A similar procedure was adopted in \citet{qi2023practical}. The product lifetime after arrival is set to $K=7$. Based on this baseline setting ($h=1$, $b=10$, $\theta=10$, $K=7$, $R=4$), we vary one parameter at a time to construct additional scenarios: $b$ and $\theta$ over $\{5,10,15,20\}$, $K$ over $\{4,5,6,7\}$, and $R$ over $\{3,4,5,6\}$. This design allows a comprehensive evaluation and robust comparison across methods.

Next, we introduce the performance metric used in our experiments. Because lead times are positive, the costs incurred during the initial periods, specifically for all $t \in [1,\cdots, t^{(r)}_1 + L^{(r)}_1 - 1]$ and for each $r \in \{1, \dots, R\}$, are identical across all policies regardless of the initial inventory configuration. Here, $t^{(r)}_1 +L^{(r)}_1$ denotes the arrival time of the first order in the $r$th simulation path. Since these initial costs do not reflect difference in policy performance, we exclude them from the performance evaluation.

For a single product, the performance of a policy $\pi$  is measured by the average cost across all periods and all $R$ simulation paths, defined as follows:
\begin{equation} \label{eq:test_measure}
\bar{C}^{\pi} = \frac{1}{R} \sum_{r=1}^{R} \left[ \frac{\sum_{t=t^{(r)}_1 + L^{(r)}_1}^{T^{\text{out}}} C_t^{\pi(r)}}{T^{\text{out}} - t^{(r)}_1 - L^{(r)}_1 + 1} \right],
\end{equation}
where $C_t^{\pi(r)}$ denotes the realized cost in period $t$ under policy $\pi$ in the $r$th simulation path.

\subsubsection{Model Setup and Hyperparameter Tuning} \label{sec:tuning}

We describe the DNN model configurations and our hyperparameter tuning procedure.

\emph{\textbf{Model setup.}} 
All DNN models are trained on a pooled dataset combining observations from all products and use the Adam optimizer. In the prediction step of the PTO-PB benchmark, we use separate long short-term memory (LSTM) networks \citep{hochreiter1997long} to forecast demand and lead time, respectively. Two embedding layers in the networks are included to encode DC and SKU indices, allowing the model to learn inter-product relationships. All LSTM models are trained using the mean squared error (MSE) loss.

For the E2E policies (as shown in Figures~\ref{fig:e2e-bb} and \ref{fig:e2e-pil}), the demand and lead time modules also utilize LSTM networks, while the order module and PIL module are implemented using multi-layer perceptrons (MLPs) with rectified linear unit (ReLU) activations. The embedding layers are also incorporated in the demand and lead time modules. The POI calculation module in the E2E-PIL policy implements the system dynamics directly and does not require parameter learning. The loss function used in the E2E models is the realized marginal cost defined in Equation~\eqref{eq:loss}, with regularization of demand, lead time and POI prediction error.

\emph{\textbf{Hyperparameter Tuning.}}
Details of the tuning procedure are provided in Appendix~\ref{ecsec:tuning}. Following  \citet{gijsbrechts2022can}, we first tune and fix the basic hyperparameters consistently across all experimental settings. The tuned configurations of these hyperparameters are reported in Table~\ref{tab:hyperparams-best}. We then tune the regularization and boosting parameters for each specific setting, using the search ranges reported in Appendix~\ref{ecsec:tuning}.
\begin{table}[htbp!]
\TABLE
{The tuned basic hyperparameters for the PTO and E2E frameworks using the real-world datasets.
    \label{tab:hyperparams-best}}
{\begin{tabular}{l 
                    >{\centering\arraybackslash}m{3.5cm} 
                    >{\centering\arraybackslash}m{2cm} 
                    >{\centering\arraybackslash}m{2cm}}
        \toprule
        \multirow{2}{*}{\textbf{Hyperparameter}}  & \multicolumn{2}{c}{\textbf{PTO }} & \multirow{2}{*}{\textbf{E2E}} \\
        \cmidrule(lr){2-3}
        & \textbf{Demand} & \textbf{Lead time} & \\
        \midrule
        Hidden size for demand module &  128 & - & 64 \\
        Hidden size for lead time module & - & 64 & 32 \\
        Hidden size for order/PIL module & - & - & 256\\
        Embedding size &  5 & 1 & 15 \\
        Number of LSTM layers &  3 & 2 &2 \\
        Weight decay &  $0$ & $10^{-4}$ & $10^{-6}$\\
        Batch size &  256 & 256 & 128\\
        Learning rate &  $10^{-2}$ & $10^{-2}$ & $10^{-3}$ \\
        Learning rate decay rate &  0.6 & 0.8 & 0.8 \\
        Learning rate decay step & 5 & 1 & 5 \\
        \bottomrule
    \end{tabular}}
{}
\end{table}

\subsubsection{Numerical Results} \label{sec:real_eval}

\begin{figure}[htbp]
     \FIGURE
    {\includegraphics[width=0.8\linewidth]{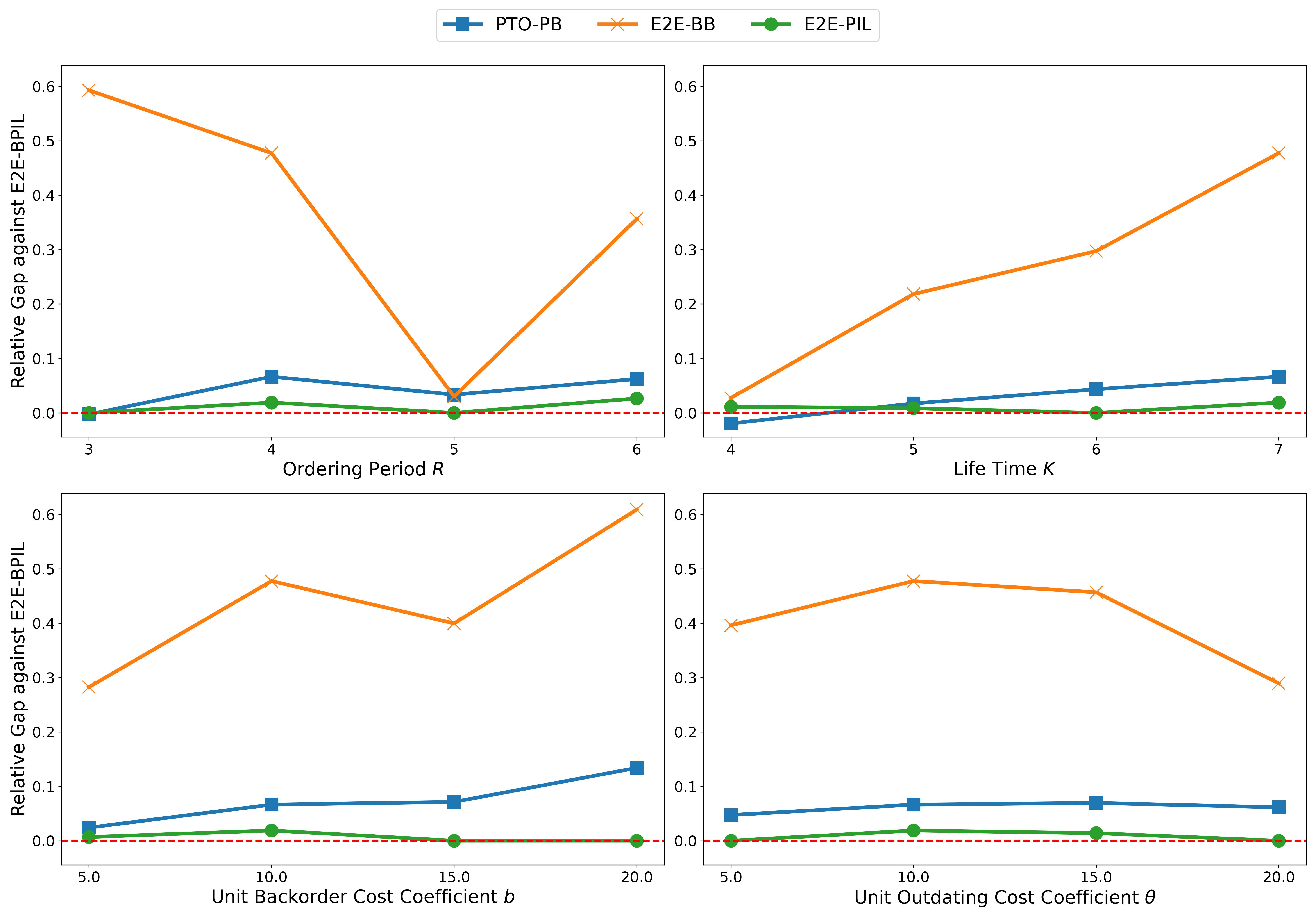}} 
{Relative gap of all other policies against $\pi^{\text{E2E-BPIL}}$ in real-world data across different problem settings. \label{fig:real_sensitivity}}
{}
\end{figure}

For each experimental configuration and method, we calculate the average cost $\bar{C}^{\pi}$ defined in Equation~\eqref{eq:test_measure} and then average  across all products. Figure~\ref{fig:real_sensitivity} presents the relative cost gap $\frac{\bar{C}^\pi - \bar{C}^{{\text{E2E-BPIL}}}}{\bar{C}^{{\text{E2E-BPIL}}}}$
of each policy~$\pi$ with respect to $\pi^{\text{E2E-BPIL}}$.

As shown in Figure~\ref{fig:real_sensitivity}, in most cases,
the cost rankings generally follow the order
$
\pi^{\text{E2E-BPIL}} < \pi^{\text{E2E-PIL}} < \pi^{\text{PTO-PB}} < \pi^{\text{E2E-BB}},
$
demonstrating the practical effectiveness of our proposed E2E-PIL policy and its boosted variant. Relative to the fully black-box E2E-BB policy, E2E-PIL embeds the PIL structure and thereby restricts the class of admissible ordering rules. This reduces the complexity of the state-to-action mapping and yields a more favorable bias–variance tradeoff, which can translate into improved out-of-sample performance. In Section~\ref{sec:explanation}, we formalize this intuition using an excess risk decomposition by comparing the hypothesis spaces induced by theory-guided and fully black-box E2E models. This perspective aligns with the ``learning less is more'' paradigm \citep{xie2024vc}: embedding domain expertise within learning models can yield more effective and data-efficient policies.

We further find that the boosted E2E-BPIL policy delivers consistent additional improvements over E2E-PIL, with larger gains in the synthetic experiments (Section~\ref{num:syn}). This pattern is consistent with the co-learning with constant boosting approach of \citet{feng2023transfer}, which improves a baseline decision rule through a shared, data-driven correction learned across related systems. In our multi-product setting, the same SKU may be replenished across multiple DCs, so the resulting SKU--DC pairs often exhibit similar demand and lead time characteristics. Consequently, the mapping from $(\bm x_{t_m},\bm z_{t_m})$ to $q_m$ shares common structure across products and locations. The boosting step in the ODA framework leverages this shared structure to deliver additional cost reductions.

However, the E2E approaches do not always dominate the PTO method. For instance, when the product lifetime is $K=4$, all E2E policies perform worse than the PTO-PB benchmark. This pattern is not unexpected for two reasons. First, PB is known to be relatively effective when lifetime is short. Under zero lead time, \citet{chao2015approximation} showed that PB admits a worst case performance bound of $(2+\frac{(K-2)h}{Kh+\theta})$, which increases monotonically from $2$ to $3$ in $K$ for $K\ge2$. This monotonicity suggests that PB becomes comparatively less favorable as lifetime grows, and therefore can be more competitive when $K$ is small. A similar pattern holds under positive lead times \citep{chao2018approximation}. Second, more generally, the literature reports regimes in which a simpler prediction-then-optimization pipeline can outperform more integrated approaches, for example when training data are limited \citep{hu2022fast} or when noise misspecification is mild \citep{elmachtoub2025dissecting}.

\subsection{Robustness Testing based on Synthetic Data} \label{num:syn}
In Section~\ref{num:real}, we compare the methods using real-world data and argue that theory-guided E2E policies can significantly reduce model complexity, ultimately leading to lower total costs. To further validate and robustly demonstrate this conclusion, we conduct an extensive comparative analysis using synthetic data that span a broad range of demand and lead time dynamics.

\emph{\textbf{Data Generation.}} To emulate real-world conditions, we generate demand, lead time, and associated feature data for $I=50$ SKUs across $J=20$ DCs over a horizon of $T=300$ days. The first $T^{\text{in}}=200$ days are used for in-sample training, and the remaining $T^{\text{out}}=100$ days are used for out-of-sample evaluation for all $I \cdot J = 1{,}000$ SKU-DC pairs. Throughout this subsection, we assume that demand follows the multivariate additive (MA) model \citep{han2023deep}. Specifically, the demand for SKU $i$ in DC $j$ during period $t$ is generated as
\begin{equation*}
D^{(i,j)}_{t} = \exp(x_{t,1}^{D(i,j)} - 0.5) + 2(x_{t,2}^{D(i)} + x_{t,3}^{D(j)} - 1)^2 + |x_{t,4}^{D} - 0.5| + \epsilon^{D(i,j)}_{t},
\quad \forall i \in [I], j \in [J], t \in [T],
\end{equation*}
where $x_{t,1}^{D(i,j)}$ is a SKU--DC specific feature,
$x_{t,2}^{D(i)}$ and $x_{t,3}^{D(j)}$ are SKU-specific and DC-specific shared features, respectively,
and $x_{t,4}^{D}$ is a global feature common to all products.
The noise term $\epsilon^{D(i,j)}_{t}$ captures unobserved factors and is specific to each SKU--DC pair. For each SKU $i$ and DC $j$, the lead time $L^{(i,j)}_{t}$ is generated as a stochastic process without covariates.

We consider four distinct data-generating configurations and generate $20$ independent instances for each configuration.
Performance is averaged across all instances.
The configurations are (see Appendix~\ref{ecsec:dgp} for details): 1) \textit{IC (Independent–Constant)}: independent and identically distributed (i.i.d.) demand, and constant lead time; 2)  \textit{CC (Correlated–Constant)}: temporally correlated demand, and constant lead time; 3) \textit{CR (Correlated–Random)}: correlated demand, and correlated random lead time; 4) \textit{SCR (Shock–Correlated–Random)}: correlated demand and correlated random lead time, which are simultaneously influenced by a common shock.

\emph{\textbf{Numerical Results.}} The model setup, hyperparameter tuning, and performance evaluation follow a similar procedure as in Section~\ref{sec:tuning}, also see Appendix~\ref{ecsec:tuning} for details. Since we have $20$ instances for each configuration, the performance is reported using mean–variance metrics or boxplots.

\begin{table}[htbp!]
\TABLE
{Average cost (standard deviation) for all the policies across four configurations.
    \label{tab:configs_cost}}
{\begin{tabular}{lcccc}
      \toprule
      \textbf{Policy/Configuration} & \textbf{IC} & \textbf{CC} & \textbf{CR} & \textbf{SCR} \\ 
      \midrule
      PTO-PB   & 9.317 (0.525) & 15.987 (1.089) & 19.480 (1.190) & 23.607 (1.318) \\ 
      E2E-BB   & 8.934 (0.544) & 16.322 (2.077) & 20.237 (2.737) & 24.890 (2.226) \\ 
      E2E-PIL  & 8.886 (0.517) & 15.055 (1.352) & 19.146 (2.232) & 22.993 (1.460) \\ 
      E2E-BPIL & 8.814 (0.487) & 14.834 (1.054) & 18.539 (1.361) & 22.234 (1.158) \\ 
      \bottomrule
    \end{tabular}}
{}
\end{table}

Table~\ref{tab:configs_cost} presents the performance of all proposed policies under the baseline problem setting (i.e., $h=1, b=10, \theta=10, K=7, R=4$) across four configurations. As the system evolves from simple i.i.d.\ demand with constant lead times to a more complex SCR setting with highly correlated demand and stochastic lead times, total costs increase for all policies. This increase reflects the greater difficulty of decision making under stronger temporal dependence and lead time uncertainty.

The cost rankings in Table~\ref{tab:configs_cost} broadly align with the findings reported in Section~\ref{sec:real_eval}. Both the E2E-PIL policy and its boosted variant consistently outperform the alternatives across all configurations. Specifically, the unboosted E2E-PIL policy reduces total cost by approximately $2\%$–$6\%$ relative to the PTO-PB policy, while the boosted E2E-BPIL policy achieves further improvements. 

\begin{figure}[htbp]
     \FIGURE
    {\includegraphics[width=0.8\linewidth]{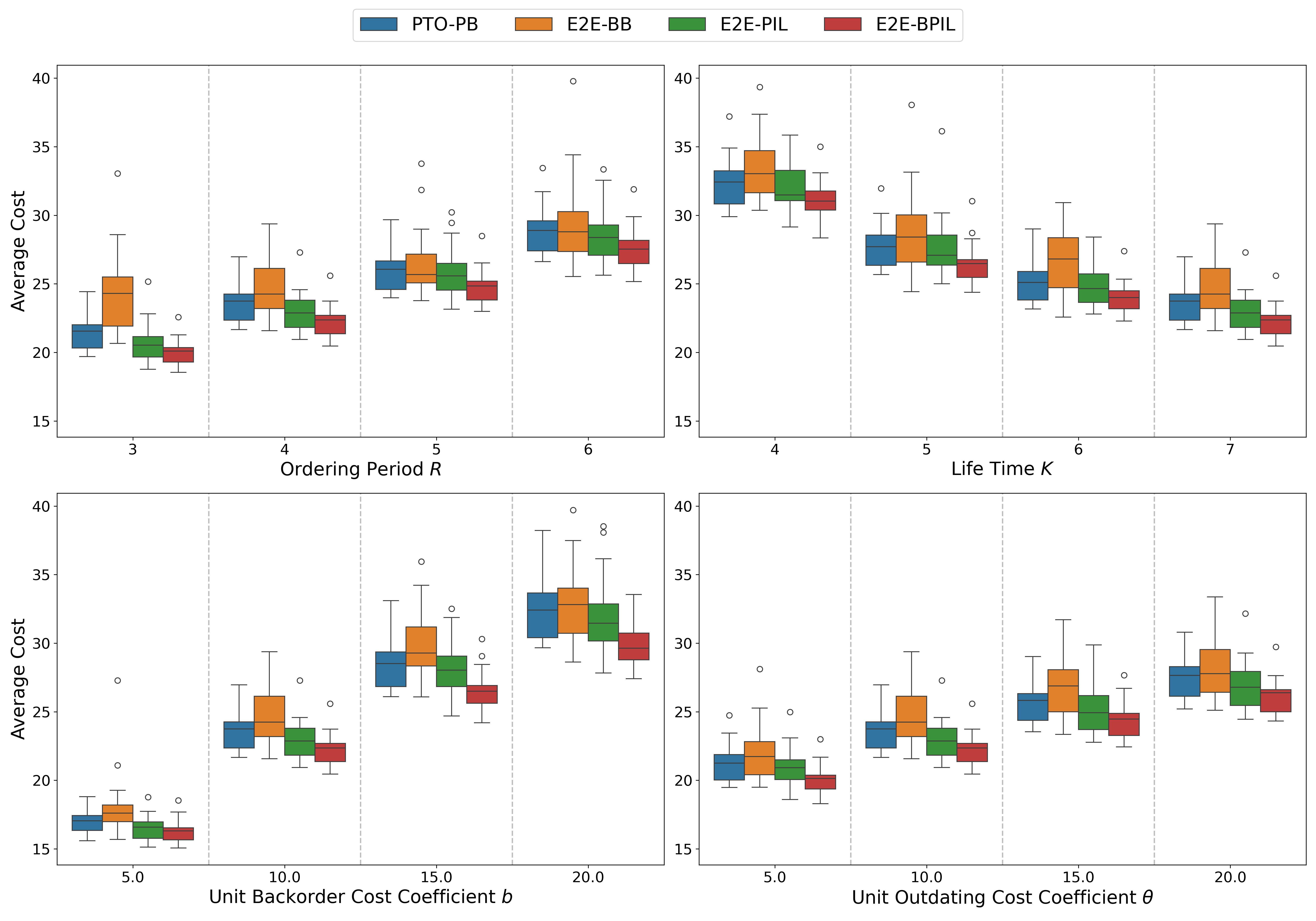}} 
{Boxplot of the average costs of all policies in synthetic data (SCR setting) across different problem settings.\label{fig:syn_sensitivity}}
{}
\end{figure}

We vary problem parameters across the same ranges as in Section~\ref{num:real} under the SCR setting, with results shown in Figure~\ref{fig:syn_sensitivity}. E2E-PIL generally attains lower costs than PTO-PB and E2E-BB, and the boosting step delivers additional reductions. Appendix~\ref{ecsec:more_results} reports supplementary analyses, including the paired t-test results, cost breakdowns (holding, backorder, and outdating costs), stockout and outdating rates, as well as additional experiments using a more competitive benchmark that empirically tunes the balancing parameter of the PB heuristic.

Overall, the results from synthetic experiments reinforce our finding that theory-guided E2E policies can substantially reduce model complexity and ultimately lower total costs. We provide further discussion of the underlying mechanisms in Section~\ref{sec:explanation}.

\section{Why E2E-PIL Outperforms E2E-BB} \label{sec:explanation}

In the previous section, using both real-world data and extensive synthetic experiments, we find that the theory-guided E2E approach is particularly effective when data are limited. This is common in inventory settings because each SKU--DC pair yields only a modest number of ordering observations and older data may be less informative as demand and lead times evolve. In this operational setting, we provide a theoretical rationale for why E2E-PIL can outperform the fully black-box E2E-BB policy under limited data.

Section~\ref{sec:excess_risk} introduces an excess risk decomposition, which serves as the foundation for our analysis. Section~\ref{sec:special_case} studies a stylized setting of nonperishable products with zero lead time and provides formal insight into the performance gap between E2E-PIL and E2E-BB. Finally, Section~\ref{sec:general_case_num} presents additional numerical results to demonstrate that the same mechanism extends to more general settings.

\subsection{Excess Risk Decomposition} \label{sec:excess_risk}

In this subsection, we formally introduce the concept of excess risk decomposition. For clarity, we slightly redefine notations used only in this section.

Let $\mathcal X$ denote the input space and $\mathcal Y$ denote the label space, and let $\mathcal P$ be the joint distribution over $\mathcal X\times\mathcal Y$. Given a loss function $\ell:\mathcal Y\times\mathcal Y\to\mathbb R_+$, the population risk of a measurable function $g:\mathcal X\to\mathcal Y$ is
$\mathcal R(g):=\mathbb E_{(x,y)\sim\mathcal P}[\ell(g(x),y)]$.
Let $\mathcal G={g:\mathcal X\to\mathcal Y}$ denote the set of measurable functions and define the population risk minimizer as $g^*\in\arg\min_{g\in\mathcal G}\mathcal R(g)$. 

We observe training data $S:=\{(x_i,y_i)\}_{i\in [n]}$ consisting of $n$ i.i.d.\ samples from $\mathcal P$ and restrict attention to a hypothesis class $\mathcal F\subseteq\mathcal G$ (e.g., linear models or a fixed DNN architecture). The empirical risk is
$\hat{\mathcal R}_n(f):=\frac{1}{n}\sum_{i=1}^n\ell(f(x_i),y_i)$,
and the empirical risk minimization (ERM) returns an optimizer $\hat f_n\in\arg\min_{f\in\mathcal F}\hat{\mathcal R}_n(f)$. Our focus is the excess risk
$\Delta\mathcal R(\hat f_n):=\mathcal R(\hat f_n)-\mathcal R(g^*)$.
The next lemma gives a standard decomposition.

\begin{lemma}  \label{lem:excess_risk} 
The excess risk of the empirical risk minimizer $\hat{f}_n$ is bounded by
    \begin{equation*}
         \Delta \mathcal{R}(\hat{f}_n)\leq\inf_{f\in \mathcal F}\bigl[\mathcal R(f)-\mathcal R(g^*)\bigr]+2\sup_{f\in\mathcal F}\bigl|\mathcal R(f)-\hat{\mathcal R}_n(f)\bigr|.    \label{eq:excess_risk}
    \end{equation*}
\end{lemma}

This lemma follows from Sections 4.1 and 4.2 of \citet{Mohri2018foundations}. Lemma~\ref{lem:excess_risk} decomposes the excess risk into two components that capture distinct sources of suboptimality. The first term, $\inf_{f\in \mathcal F}\bigl[\mathcal R(f)-\mathcal R(g^*)\bigr]$, is the \emph{approximation error}.  It measures the best achievable gap between $\mathcal F$ and $g^*$, and is driven by the expressive limits of the hypothesis class of $\mathcal F$. When $g^*\notin \mathcal F$, this term remains strictly positive regardless of sample size. The second term, $\sup_{f\in\mathcal F}\bigl|\mathcal R(f)-\hat{\mathcal R}_n(f)\bigr|$, is the \emph{generalization error}. It quantifies the maximal deviation between empirical and population risks over $\mathcal F$, capturing the statistical uncertainty induced by finite samples.

Together, in the same spirit as the usual bias–variance trade-off, these terms highlight an excess risk trade-off induced by the choice of $\mathcal F$. Enlarging $\mathcal F$ can reduce the approximation error, but it typically increases the generalization error because a richer class is harder to learn reliably from limited data. Conversely, a more restrictive class may generalize better but can incur larger approximation error. This decomposition motivates incorporating human knowledge from long-standing inventory theory to impose structure-preserving restrictions when designing $\mathcal F$, so as to better balance these effects.

\subsection{A Special Case: Nonperishable Products and Zero Lead Time } \label{sec:special_case}

In this subsection, we study a special case in which the product lifetime approaches infinity ($K \rightarrow \infty$) and the lead time is zero (i.e., $\bar{L}=0$). This yields a nonperishable, zero-lead-time system in which the inventory state reduces to a scalar $z\in\mathbb R$. We show that, in this setting, using inventory theory to parameterize a base-stock decision, rather than directly learning an order-quantity mapping, can reduce the generalization error without increasing the approximation error.

We cast this setting in a supervised learning framework. The label is demand $D_t\in \mathbb R$, the inputs are the demand feature vector $\bm X^D_t\in \mathbb R^p$ (denoted by $\bm X_t$ in this subsection) and the inventory state $z_t\in \mathbb R$. Demand is generated as $D_t=f^D(\bm X_t)+\epsilon_t$, consistent with Section~\ref{sec:Model}. We assume that $\bm X_t$ and $\epsilon_t$ are i.i.d.\ over time and that demand is bounded, $D_t\in[0,\bar{D}]$ for all $t$. An order-quantity policy $f:\mathbb R^p\times\mathbb R\rightarrow \mathbb R$ maps $(\bm X_t,z_t)$ to a replenishment decision.  By Proposition 1 in \citet{ding2024feature}, the myopic base-stock policy that raises the inventory level to minimize the single-period Newsvendor cost is optimal for this dynamic setting. Accordingly, for a realized input-label pair $(\bm x,z,d)$, we define the loss as $\ell(f(\bm x,z),d)=b[d-f(\bm x,z)-z]^++h[f(\bm x,z)+z-d]^+$, and the population risk equals the expected Newsvendor cost $\mathbb E_{\bm X,z,D}[b(D-f(\bm X,z)-z)^++h(f(\bm X,z)+z-D)^+]$. While we treat $z$ as random in the expectation, the arguments below can also be interpreted conditional on any fixed $z$.

It can be shown that the globally optimal policy is $g^*(\bm x,z)=f^D(\bm x)+F_\epsilon^{-1}(\tfrac{b}{b+h})-z $, where $F_\epsilon^{-1}$ denotes the inverse cumulative distribution function of $\epsilon$. We allow the order quantity to be negative, which can be interpreted as a reverse inventory adjustment (e.g., disposal or returns), as is common in inventory models that include explicit disposition or reverse flows \citep{HoadleyHeyman1977,YangZhang2014}. Such reverse flows also arise in practice. For example, in community group buying, unsold items may be returned upstream to suppliers after fulfillment on a daily basis \citep{PengLiRongLuoMaZhao2025}.

In practice, we only observe historical data $\{(\bm{x}_i, z_i, d_i)\}_{i=1}^n$ and approximate the optimal policy within a hypothesis space $\mathcal{F} \subseteq \{ f: \mathbb{R}^p \times \mathbb{R} \to \mathbb{R} \}$. As is standard, we assume $\mathcal F$ is uniformly bounded, i.e., $\|f\|_\infty\leq Q$, for all $f\in \mathcal F$ and some positive constant $Q$ since the output can always be clipped. We also assume $\mathcal F$ is closed under pointwise convergence (as is typical for parametric DNN or polynomial classes), and thus compact. To facilitate the analysis, we further impose the following assumption.

\begin{assumption}\label{asp:hypothesis}
The hypothesis space $\mathcal F$ is closed under the following ``fix-and-shift'' transformation: for any $f\in\mathcal F$, its extended space $\{\,f_c:\mathbb R^p\times \mathbb R\to \mathbb R \mid \forall (\bm x,z)\in\mathbb{R}^p \times \mathbb{R}, f_c(\bm x,z)=f(\bm x,c)+c-z,\, c\in\mathbb R\,\}\subseteq\mathcal F$.
\end{assumption}

To interpret Assumption~\ref{asp:hypothesis}, consider an example of a one-dimensional feature $x$, where the optimal policy is given by $g^*(x,z)=999x-z$. Suppose a candidate $f(x,z)=x\cdot \max(z,1000)-z$ belongs to $\mathcal F$. Assumption~\ref{asp:hypothesis} then implies that, for any constant $c\in\mathbb R$, the transformed function $f_c(x,z)=f(x,c)+c-z=x\cdot \max(c,1000)-z$ also lies in $\mathcal F$. In this example, $f$ cannot be best in class because a better approximation is obtained by choosing $c=1000$, namely $f_{1000}(x,z)=1000x-z$. Indeed, for any $(x,z)$,
$$
|f_{1000}(x,z)-g^*(x, z)|=|x|\leq |x\cdot \max(z-1000,1)|=|f(x,z)-g^*(x, z)|.
$$
Thus, the extended space contains a function that is uniformly closer to the optimum. In the proof of Proposition~\ref{prop:cons_H}, we formalize this intuition by showing that for any $f\in\mathcal F$, the population risk can be weakly improved within its fix-and-shift family.

Assumption~\ref{asp:hypothesis} is mild. Next, we show that two commonly used hypothesis classes satisfy it. We begin by defining the polynomial and DNN function spaces.

\begin{definition}
The $\mathcal A$-th order polynomial function space is defined by
$$
\mathcal F^{\text{poly}}_{\mathcal A}
:=\left\{\, f:\mathbb R^p\times\mathbb R\rightarrow \mathbb R \,\middle |\,
f(\bm x,z)
=\sum_{0\le i_1+\cdots+i_p+j\le \mathcal A}
\left(\alpha_{i_1,\ldots,i_{p},j}\,
z^{j}\prod_{k=1}^{p} x_k^{i_k}\right)
,\,
\alpha_{i_1,\ldots,i_{p},j}\in\mathbb R
\right\},
$$
where $\{\alpha_{i_1,\dots,i_p,j}\}$ are the coefficient parameters of the corresponding monomials.
\end{definition}

\begin{definition}
The linear-augmented DNN space of depth $\mathcal A$ with layer widths $\{w_0,\dots,w_{\mathcal A+1}\}$, where $w_0=p+1$ is the input dimension and $w_{\mathcal A+1}=1$ is the output dimension, is defined by
\begin{align*}
    \mathcal F^{\text{DNN}}_{\mathcal A,\bm w}
:=\Bigl\{\,f:\mathbb R^p\times\mathbb R\rightarrow \mathbb R \,\Big |\, &
f(\bm x,z)
=l^{\text{aug}}((\bm x;z))+l_\mathcal A\circ \sigma\circ \cdots \circ \sigma\circ l_0((\bm x;z))
,\\
& \bm W^{\text{aug}}\in\mathbb R^{w_{0}},\;
\bm W_i\in\mathbb R^{w_{i+1}\times w_i},\;
\bm b_i\in\mathbb R^{w_{i+1}}
\Bigr\},
\end{align*}
where $(\bm x;z)$ denotes the concatenated input of dimension $p+1$, $l^{\text{aug}}(\bm x)=\bm W^{\text{aug}\top}\bm x$ denotes the linear connection layer directly linking the input and output layer, $l_i(\bm x)=\bm W_i\bm x+\bm b_i$ denotes the hidden layer linear connection for $i=0,\dots,\mathcal A$, and $\sigma$ denotes the activation function (e.g., ReLU).
\end{definition}

\begin{lemma}\label{lem:poly_dnn} For any $\mathcal A\geq 1$ and $w_i\geq 1, i=1,\dots,\mathcal A$, 
both $\mathcal F^{\text{poly}}_{\mathcal A}$ and $\mathcal F^{\text{DNN}}_{\mathcal A,\bm w}$ satisfy the Assumption~\ref{asp:hypothesis}.
\end{lemma}

A practitioner without prior inventory knowledge may attempt to learn the full mapping in $\mathcal F$. In this special case, however, the optimal policy is known to have a base-stock form, which fixes how the inventory state $z$ enters the decision. This suggests designing the hypothesis space to avoid relearning the dependence on $z$. To do so, we define the following reduced function space:
\begin{equation}\label{eq:reduced_V}
\mathcal V
:=\Bigl\{\, v:\mathbb R^p\to\mathbb R \ \Bigm|\ 
\exists f\in\mathcal F \text{ such that } 
f(\bm x,z)=v(\bm x)-z,\ \forall(\bm x,z)\in\mathbb R^p\times\mathbb R
\,\Bigr\}.
\end{equation}
Under Assumption~\ref{asp:hypothesis}, $\mathcal V$ is not empty by letting $v(\bm x)=f(\bm x,c)+c$ for any $f\in\mathcal F$ and $c\in \mathbb R$.
We then define the constrained space where the effect of $z$ input is fixed:
\begin{equation}\label{eq:cons_H}
\mathcal F'
:=\Bigl\{\, f:\mathbb R^p\times\mathbb R\to\mathbb R \ \Bigm|\ 
\forall(\bm x,z)\in\mathbb R^p\times\mathbb R,\,f(\bm x,z)=v(\bm x)-z,\, v\in\mathcal V
\,\Bigr\}.
\end{equation}
In $\mathcal F'$, it suffices to learn the reduced function $v$, simplifying learning. The proposition below formally shows that $\mathcal F'$ yields a smaller generalization error than $\mathcal F$, while keeping the same approximation error.

\begin{proposition} \label{prop:cons_H}
 In the non-perishable multi-period inventory system with zero lead time and stationary demand, with Assumption~\ref{asp:hypothesis}, for the original hypothesis space $\mathcal F$ and the constrained space $\mathcal F'$ defined in Equation~\eqref{eq:cons_H}, we have
    \begin{subequations}  
    \label{eq:prop_risks}
        \begin{align}
            \inf_{f\in \mathcal F'}\bigl[\mathcal R(f)-\mathcal R(g^*)\bigr]&=\inf_{f\in \mathcal F}\bigl[\mathcal R(f)-\mathcal R(g^*)\bigr],\\
            \sup_{f\in\mathcal F'}\bigl|\mathcal R(f)-\hat{\mathcal R}_n(f)\bigr|&\leq \sup_{f\in\mathcal F}\bigl|\mathcal R(f)-\hat{\mathcal R}_n(f)\bigr|.
        \end{align}
    \end{subequations}
\end{proposition}

Proposition~\ref{prop:cons_H} suggests a simple message: when the policy structure is known, we can restrict attention to a lower-dimensional decision rule without worsening approximation performance, while potentially improving generalization. We next illustrate the magnitude of this improvement in a concrete example based on a polynomial hypothesis space.

\begin{example}
  Consider a one-dimensional feature $x$ and an $\mathcal A$-th order polynomial hypothesis class for order quantity mapping:
  $$
    \mathcal F
      := \left\{\,
         f:\mathbb R\times\mathbb R\rightarrow \mathbb R \,\middle |\,f(x,z)=\sum_{0 \leq i+j \leq \mathcal A} \alpha_{i,j} x^i z^j\, , \,  \alpha_{i,j} \in \mathbb R \right\}.
  $$
  The constrained class $\mathcal F'$ in Equation~\eqref{eq:cons_H} fixes the dependence on $z$, and therefore reduces to
  $$
    \mathcal F'
      := \left\{\, f:\mathbb R\times\mathbb R\rightarrow \mathbb R \,\middle |\,f(x,z)=\sum_{k=0}^\mathcal A \alpha_k x^k - z\,  ,  \, \alpha_k \in \mathbb R \right\}.
  $$
  By Theorem 11.6 in \citet{Mohri2018foundations}, the pseudo–dimension of $\mathcal F$ is $\frac{(\mathcal A+2)(\mathcal A+1)}{2}$ and that of $\mathcal F'$ is $\mathcal A+1$, treating all high-order and interaction terms as separate inputs. Moreover, within either class, the loss is uniformly bounded by $b\bar{D}+hQ$. Therefore, by Theorem 11.8 in \citet{Mohri2018foundations}, for any $\delta>0$, with probability $1-\delta$, the generalization error of a hypothesis space $\mathcal H$ with pseudo–dimension $\mathscr{D}$ is bounded by
  \begin{equation*}
    \sup_{f\in\mathcal H}[\mathcal R(f)-\hat{\mathcal R}_n(f)]\leq 
    (b\bar{D}+hQ)\left[\sqrt{\frac{2\mathscr{D}\log\frac{en}{\mathscr{D}}}{n}}+\sqrt{\frac{\log\frac{1}{\delta}}{2n}}\right].
  \end{equation*}

  That is, the generalization error scales as $O(\sqrt{\frac{\mathscr{D}\log\frac{n}{\mathscr{D}}}{n}})$. Consequently, to obtain a comparable generalization bound, the sample size required by $\mathcal F'$ is approximately $\frac{2}{2+\mathcal A}$ times that required by $\mathcal F$. This reduction becomes more pronounced as $\mathcal A$ increases, reflecting the larger complexity gap between the two classes.
\end{example}

\subsection{The General Cases} \label{sec:general_case_num}

We next move beyond the special case and consider more general inventory settings. For perishable systems with random lead times, the structure of the optimal policy is not fully characterized. In such settings, imposing additional structural constraints on a rich hypothesis class can, in principle, increase approximation error by excluding the globally optimal mapping. However, high-performing heuristics for perishable inventory, such as the PIL policy, have strong theoretical support and good empirical performance. This suggests that embedding the PIL structure may entail only a modest loss in optimality while offering substantial statistical benefits in finite samples. In particular, the E2E-PIL architecture uses a POI calculation module to fix how on-hand inventory $\bm z$ enters the ordering decision. When the original hypothesis class $\mathcal F$ is a highly flexible DNN, this restriction can substantially reduce the effective complexity of the learned state-to-action mapping, which is especially valuable when data are limited. This provides another instance of the ``learning less is more'' perspective \citep{xie2024vc}: in complex operational learning problems, incorporating a well-specified, policy-relevant structure can reduce learning burden while preserving strong decision quality.

\begin{figure}[htbp]
     \FIGURE
    {\includegraphics[width=1\linewidth]{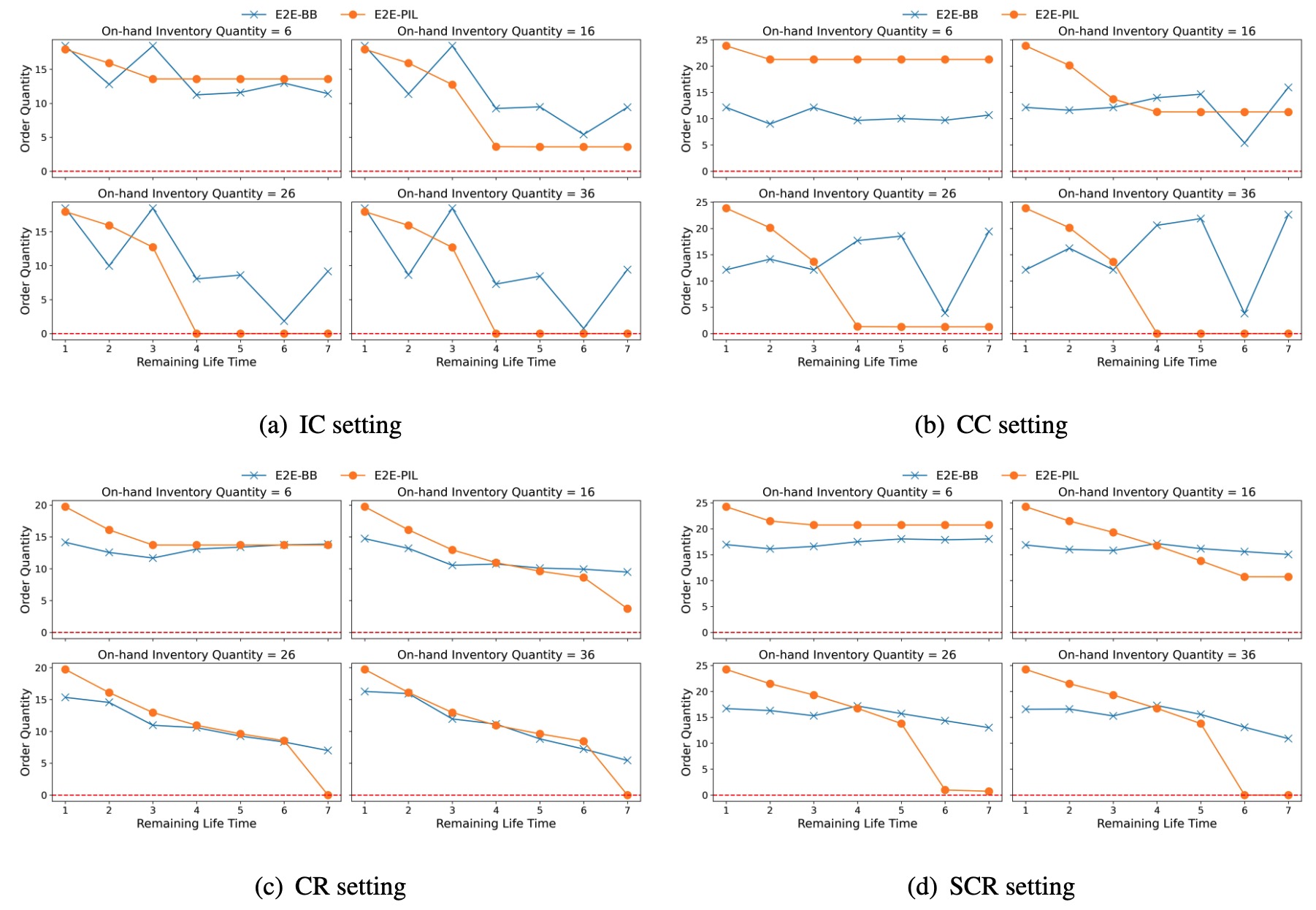}} 
{The change of order quantities with remaining life time for both the E2E-BB (blue line, cross marker) and E2E-PIL (orange line, point marker) policy. \label{fig:trained_map}}
{}
\end{figure}

We next open the black box of the two E2E policies to illustrate the advantage of E2E-PIL, which aligns with the idea of “learning less”. For ease of visualization and to facilitate an apples-to-apples comparison of the learned state-to-action mappings, we consider a controlled numerical setting. We fix the feature input $\bm{x}$ to the test data from day~50 for the first product in the first synthetic instance under each of the four configurations. Both models are trained on the same instance, as described in Section~\ref{sec:num}. To examine how each model maps the inventory state vector $\bm{z}$ to an ordering decision, we construct inputs in which exactly one of the first $K$ components of $\bm{z}$ is set to a positive value and all remaining components, including the final $\bar{L}-1$ pipeline entries, are set to zero. By varying this positive value and its associated remaining lifetime, we trace out the learned state-to-action responses. The results are reported in Figure~\ref{fig:trained_map}.

Classical inventory theory has long characterized the sensitivity of the optimal policy for perishable inventory systems under i.i.d.\ demand \citep{nahmias1975optimal, fries1975optimal}. In particular, the marginal effect of inventory is nonpositive and ordered across ages:
\begin{equation*} \label{eq:optimal_sensitivity}
-1 \leq \frac{\partial q_t^*}{\partial z_{t,K-1}} \leq
\frac{\partial q_t^*}{\partial z_{t,K-2}} \leq \cdots
\leq \frac{\partial q_t^*}{\partial z_{t,1}} \leq 0.
\end{equation*}
Figure~\ref{fig:trained_map} shows that \textsc{E2E-PIL} closely aligns with this economically meaningful monotone structure across all four configurations: the order quantity is nonincreasing in remaining lifetime, and higher on-hand inventory shifts the curve downward. The plateau at large remaining lifetimes has a natural interpretation: when inventory is sufficiently ``fresh'', it will be depleted before expiring, so additional lifetime has little marginal value for replenishment. 

In contrast, E2E-BB exhibits two complementary finite-sample pathologies. In the simpler IC and CC settings, its response is often highly irregular and non-monotone, occasionally ordering more at longer remaining lifetimes, which is consistent with overfitting to spurious correlations induced by trajectory noise and sparse state visitation. In the more complex CR and SCR settings, the same data limitation manifests differently: the E2E-BB curves become comparatively flat and less separated across on-hand levels, indicating shrinkage toward an ``average'' action when the relevant dependence on $z$ is weakly identified from limited trajectories. One plausible mechanism is that as demand and lead time dynamics become richer, the feature vector $\bm x_{t_m}$ carries stronger predictive signal; a black-box network can then rely heavily on $\bm x_{t_m}$ and treat $\bm z_{t_m}$ as secondary, producing the muted inventory-state sensitivity observed in CR and SCR. Taken together, Figure~\ref{fig:trained_map} suggests that structural guidance in E2E-PIL mitigates both local overfitting (instability) and global under-learning (insensitivity) by reducing the effective complexity of the state-to-action map.

\section{Conclusions} \label{sec:conclusion}

In this paper, we study a multi-period perishable inventory system with stochastic lead times in a data-driven setting where demand and lead time processes are unknown. We develop end-to-end policies that translate established inventory-theoretic insights into learnable neural decision rules. Across extensive numerical experiments using a real-world beverage dataset and a wide range of synthetic environments, the theory-guided E2E-PIL policy and its boosted variant, E2E-BPIL, consistently outperform the black-box E2E-BB policy and a competitive prediction-then-optimization benchmark. We further rationalize this advantage through an excess risk lens: embedding the PIL structure reduces the effective complexity of the learned policy class, improving learning efficiency in limited-data regimes while largely preserving approximation flexibility. Overall, our results highlight a concrete role for human knowledge: inventory-theoretic structure can be encoded as a theory-guided structural constraint in end-to-end learning, yielding more sample-efficient and reliable policies in complex perishable systems.

This work suggests several directions for future research. On the modeling side, it would be valuable to extend the framework to broader inventory settings whenever theory provides useful policy structure, including systems with capacity constraints, setup costs, multi-echelon networks, and interactions across items (e.g., substitution or shared resources). An important related question is how far theory-based structure can go across different operational contexts. Reinforcement learning is another promising direction, but standard methods are often data-hungry and can be unstable in inventory settings where each product yields only limited observations. Our policy-embedding approach offers a potential remedy by restricting learning to a structured, theory-guided policy class, thereby improving sample efficiency.

% Acknowledgments here
%\ACKNOWLEDGMENT{The authors gratefully thank the reviewers of POM.}

%%REFERENCES%%
%%%%%%%%%%%%%%%%%%%%%%%%%%%%%%%%%%%%%%%%%%%%%%%%%%%%%%%%%%%%%%%%%%%%%%%%%%%%%%%%%%%%%%%%%%%%%%%%%%%%%%%%%%%%%%%%%%%%%%%%%%%%%%%%%%%%
%% This template complies references using bibtex. You will need to use pomsref.bst file for biblography style.
%REFERENCES USING BIBTEX FILES
%%%%%%%%%%%%%%%%%%%%%%%%%%%%%%%%%%%%%%%%%%%%%%%%%%%%%%%%%%%%%%%%%%%%%%%%%%%%%%%%%%%%%%%%%%%%%%%%%%%%%%%%%%%%%%%%%%%%%%%%%%%%%%%%%%%%

\bibliographystyle{informs2014} % outcomment this and next line in Case 1
\bibliography{ref1} 

%%%%%%%%%%%%%%%%%%%%%%%%%%%%%%%%%%%%%%%%%%%%%%%%%%%%%%%%%%%%%%%%%%%%%%%%%%%%%%%%%%%%%%%%%%%%%%%%%%%%%%%%%%%%%%%%%%%%%%%%%%%%%%%%%%%%

%Hayes, R. H., G. P. Pisano. 1996. Manufacturing strategy: At the intersection of two paradigm shifts. Production and Operations Management, 5 (1), 25-41.

%%%%%%%%%%%%%%%%%%%%%%%%%%%%%%%%%%%%%%%%%%%%%%%%%%%%%%%%%%%%%%%%%%%%%%%%%%%%%%%%%%%%%%%%%%%%%%%%%%%%%%%%%%%%%%%%%%%%%%%%%%%%%%%%%%%%
%% %If you don't use BiBTex, you can manually itemize references as shown in the referneces for the electronic comapanion. See below.
 %%%%%%%%%%%%%%%%%%%%%%%%%%%%%%%%%%%%%%%%%%%%%%%%%%%%%%%%%%%%%%%%%%%%%%%%%%%%%%%%%%%%%%%%%%%%%%%%%%%%%%%%%%%%%%%%%%%%%%%%%%%%%%%%%%%%

%%%%%%%%%%%%%%%%%
\clearpage
\ECSwitch
%\clearpage

\begin{APPENDIX}{}

\noindent The Appendix contains additional materials that complement the main text. Specifically, in Section~\ref{ecsec:data_arrange}, we describe the detailed procedure for organizing the historical data into training samples; in Section~\ref{ecsec:POI_cal}, we outline the implementation of the POI calculation module; in Section~\ref{ecsec:benchmark_pb}, we present the implementation details of the PTO-PB benchmark; in Section~\ref{ecsec:descriptive}, we provide a comprehensive descriptive analysis of the real-world dataset; in Section~\ref{ecsec:tuning}, we summarize the procedure and results of the hyperparameter tuning; in Section~\ref{ecsec:dgp}, we specify the configurations of the data-generating process; in Section~\ref{ecsec:more_results}, we present additional numerical results to support our findings; and finally, in Section~\ref{ecsec:proofs}, we provide the proofs for the theoretical results.

\section{Organize the Historical Data into Training Samples} \label{ecsec:data_arrange}

We use an example to illustrate how to organize historical data into training samples in Figure~\ref{fig:organize}. Let the history window be $4$, and $K+\bar L=3$ which leads to the future demand window $3$. Assume historical records span times from $-200$ to $0$, and the history ordering points are ${-200,-198,\dots,0}$. The first usable training point is $t=-196$. For this point extract demand and feature observations from $-200$ to $-197$ as the input sequence, and extract demand observations from $-196$ to $-194$ as the target sequence. The lead time data use the same procedure, except the future lead time window is always $2$, regardless of $K+\bar L$, and the available historical lead time observations are effectively only $2$ because of missing entries at non ordering times. Once both the history and future sequences are complete, repeat the same extraction for $t=-194,-192,\dots$ to assemble the full training set. This procedure converts both input and target into fixed length sequences, making them suitable for neural network training, and as in the main text we denote the inputs for demand and lead time jointly by $\bm x\in\mathbb R^p$ for subsequent analysis.

\begin{figure}[htbp]
     \FIGURE
    {\includegraphics[width=1\linewidth]{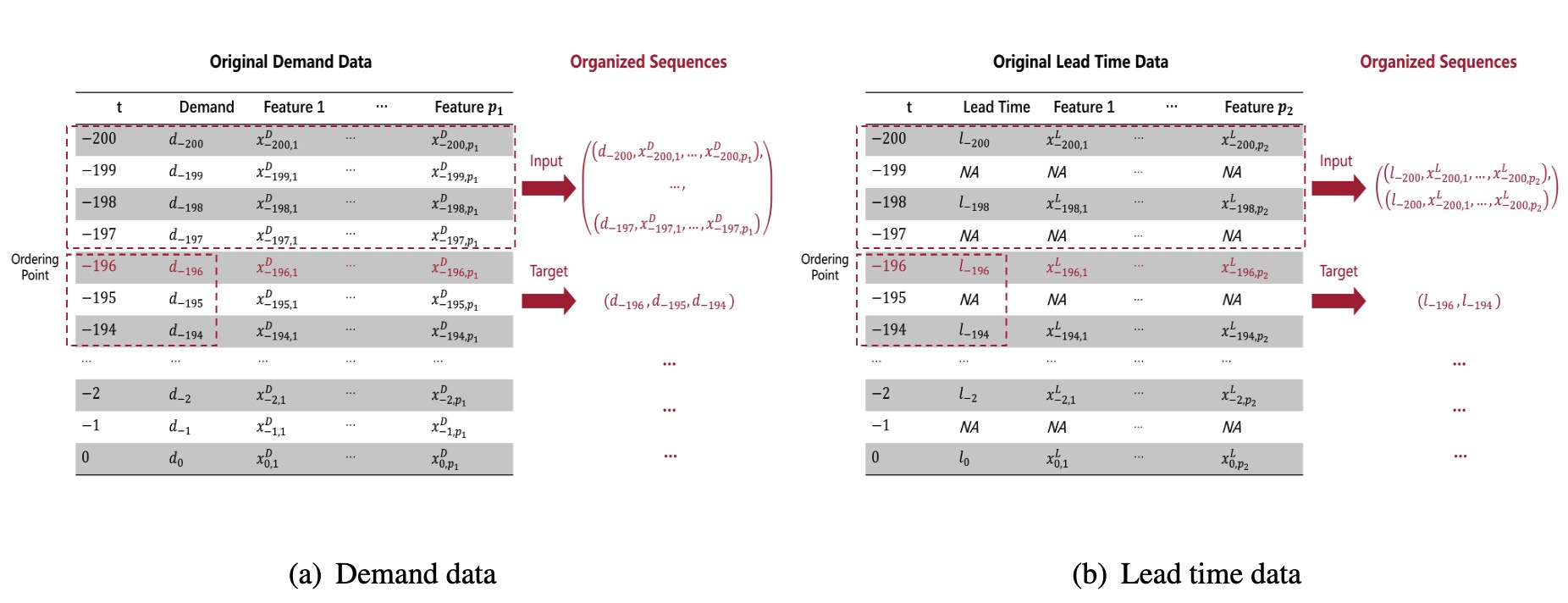}} 
{An example of organizing the historical data into training samples. \label{fig:organize}}
{}
\end{figure}

\section{Details of the POI Calculation Module} \label{ecsec:POI_cal}

We detail the POI calculation module in Algorithm~\ref{alg:POI_cal}. We first restate the exact POI calculation in subsection~\ref{sec:pil}:
\begin{equation*}\label{eq:exact_POI}
z^{\text{POI}}_{s}=\sum_{i=1}^{K+\bar L-1}z_{t_m,i}-D_{[t_m,s-1]}-B_{[t_m,s-1]}(\bm z_{t_m}),
\quad\forall s=v_m,\dots,v_m+K-1.
\end{equation*}
Based on the definition of unmet demand $\tilde{D}_s(\bm z_{t_m})$ in Equation~\eqref{eq:d_tilde}, one can verify that
\begin{equation*}\label{eq:poi_unmet_relation}
z^{\text{POI}}_s \;=\; -\bigl[\tilde{D}_s(\bm z_{t_m}) - D_s\bigr]
\;=\; D_s-\tilde{D}_s(\bm z_{t_m}).
\end{equation*}
Therefore, in the first part of Algorithm~\ref{alg:POI_cal}, once $\tilde D_s(\bm z_{t_m})$ is available, we could then compute the exact $z^{\mathrm{POI}}_s$. However, the required alignment from the arrival point to the perishing point depends on the lead time $L$ through an index selection, which is not differentiable when $L$ is an input predicted by an upstream forcasting module of the network.

The second part of Algorithm~\ref{alg:POI_cal} addresses this issue by replacing the hard index operation with a differentiable Gaussian-kernel smoothing. Specifically, when $s'=0$, $v_m$ is exactly the arrival time of the new order. The Gaussian kernel $\tilde\kappa(v_m,t_m+i)$ therefore assigns larger weights to indices $t_m+i$ closer to $v_m$, and the smoothed POI is a weighted average of the exact POI values $\bigl[D_i-\tilde D_i(\bm z_{t_m})\bigr]$ around this arrival index. The cases $s'=1,\dots,K-1$ are analogous, with the kernel center shifted to $v_m+s'$ accordingly. When $w\to 0^+$, the weights concentrate on the aligned index and the smoothing approaches the original index operation. In our numerical experiments, we set $w=0.3$ and find the approximation performs reliably across diverse scenarios. To guard against numerical issues from vanishing kernel values, one may enforce the lead time input $L$ to lie in its modeled range (e.g., $L\in[0,\bar L]$) by introducing either an out-of-range penalty or a prediction regularization term on the lead time outputs.

\begin{algorithm}[htbp]
\footnotesize
\caption{The POI Calculation Module}
\label{alg:POI_cal}
\KwIn{Future demands $\{D_t\}_{t=t_m}^{t_m+\bar L+K-1}$; lead time $L_m$; inventory state $\bm z_{t_m}$; bandwidth $w>0$.}
\KwOut{Sequence of smoothed future POIs $\{z_t^{\text{POI}}\}_{t=v_m}^{v_m+K-1}$ from the arrival point to the perishing point.}

\BlankLine
\textbf{(1) Compute $\{\tilde D_s(\bm z_{t_m})\}_{s=t_m}^{t_m+K+\bar L-1}$ (needed for exact POIs)}\;
\For{$s=0,\dots,K+\bar{L}-2$}{
Recursively compute the outdating quantity $\tilde{B}_{[t_m,t_m+s]}(\bm z)$ based on Equation~\eqref{eq:outdata_quantity}\;
}
\For{$s=0,\dots,K+\bar{L}-1$}{
Compute the unmet demand $\tilde{D}_{t_m+s}(\bm z)$ based on Equation~\eqref{eq:d_tilde}\;
}

\BlankLine
\textbf{(2) Differentiable alignment via Gaussian kernel smoothing}\;
\For{$s'=0,\dots,K-1$}{
\For{$i=0,\dots,K+\bar{L}-1$}{
$\kappa(v_m+s',t_m+i)\leftarrow \exp\!\bigl(-\frac{(L+s'-i)^2}{w}\bigr)$\;
}
\For{$i=0,\dots,K+\bar{L}-1$}{
$\tilde{\kappa}(v_m+s',t_m+i)\leftarrow \dfrac{\kappa(v_m+s',t_m+i)}{\sum_{j=0}^{K+\bar{L}-1}\kappa(v_m+s',t_m+j)}$\;
}
$z^{\text{POI}}_{v_m+s'} \leftarrow \sum_{i=0}^{K+\bar{L}-1}\tilde{\kappa}(v_m+s',t_m+i)\cdot \bigl[D_{i}-\tilde{D}_i(\bm z)\bigr]$\;
}
\KwRet Arrival-point POI $z^{\text{POI}}_{v_m}$ (for PIL ordering) and the subsequent POIs $\{z_t^{\text{POI}}\}_{t=v_m+1}^{v_m+K-1}$ (for prediction regularization).
\end{algorithm}

\section{Implementation of the PTO-PB benchmark} 
\label{ecsec:benchmark_pb}

We detail the PTO-PB benchmark here. The PTO framework consists of two main steps. First, an off-the-shelf ML model is used to predict the distributions of future demands and lead times. Then, based on these predicted distributions, heuristic algorithms determine the ordering decisions for the downstream perishable inventory system.

The prediction step follows a standard machine learning procedure. Specifically, for each order $m$, we employ an LSTM network to output estimations of the conditional mean of the future demands $\hat{\bm d}_{t_m}$ and the future lead times $\hat{\bm l}_{m}$. According to Appendix \ref{ecsec:descriptive}, we find that there is no general correlation between demand and lead time in the real-world dataset. Therefore, we use their respective historical sequences as separate inputs for each model. If such correlation does exist (e.g., the SCR setting we use in the synthetic data generation), we use both historical sequences as inputs to the models to capture the correlation. The sequence lengths are consistent with their corresponding targets (i.e., $K+\bar{L}$ for demand $\bm d_{t_m}$ and $2$ for lead time $\bm l_{m}$, as discussed in Section~\ref{sec:Model}). 

After obtaining the trained model, we compute the in-sample residuals $\hat{\bm\epsilon}_{t_m} = \bm d_{t_m} - \hat{\bm d}_{t_m}$ for demand and $\hat{\bm\eta}_{m} = \bm l_{m} - \hat{\bm l}_{m}$ for lead time, and fit them to predetermined distributions. Throughout this paper, we model these residuals with multivariate normal distributions, which capture temporal correlations within the sequences. The sampled residuals are truncated between the historical minimum and maximum values. For integer-valued lead times, the sampled continuous values are rounded down. 

To approximate the conditional distributions of future demands and lead times, we generate residual samples $\hat{\bm\epsilon}_{t_m}^{(\text{sample})}$ and $\hat{\bm\eta}_{m}^{(\text{sample})}$ from the fitted distributions, and combine them with the predicted means:
$$
\hat{\bm d}_{t_m}^{(\text{sample})} = \hat{\bm d}_{t_m} + \hat{\bm\epsilon}_{t_m}^{(\text{sample})}, 
\qquad 
\hat{\bm l}_{m}^{(\text{sample})} = \hat{\bm l}_{m} + \hat{\bm\eta}_{m}^{(\text{sample})}.
$$
The resulting conditional distributions are denoted by $\hat{F}_{\bm D|\bm x}$ for demand and $\hat{F}_{\bm L|\bm x}$ for lead time.

We use the PB algorithm \citep{chao2015approximation,chao2018approximation} in the optimization step. Leveraging the marginal cost accounting scheme, the PB policy balances the expected marginal holding and outdating costs with the marginal backorder cost:
\begin{equation} \label{eq:pb}
    \beta \mathbb{E}[H(q_{m}|\bm{z}_{t_m})+\Theta(q_{m}|\bm{z}_{t_m})|\bm x_{t_m}]=\mathbb{E}[\Pi(q_{m}|\bm{z}_{t_m})|\bm x_{t_m}],
\end{equation}
where
\begin{equation} \label{eq:balancing_coeff}
    \beta =
    \begin{cases}
    1,& K=1; \\
    \frac{Kh+\theta}{2(K+L_m-1)h+\theta},&  K>1.
    \end{cases}
\end{equation}
The expectations of both sides are taken on the conditional distributions of the future demands and lead times, which are unknown to our problem. Thus, under the PTO framework, we plug the ML-based predicted distribution $\hat{F}_{\bm D|\bm x_{t_m}}$ and $\hat{F}_{\bm L|\bm x_{t_m}}$ to get the PTO-PB policy $\pi^{\text{PTO-PB}}$. Moreover, we replace $L_m$ in Equation~\eqref{eq:balancing_coeff} by the sample conditional mean of lead time to make the balancing coefficient a known value. That is
\begin{equation*}
    \pi^{\text{PTO-PB}}(\bm x_{t_m},\bm z_{t_m})=q,\text{ s.t. } q \text{ solves Equation~\eqref{eq:pb}, with expectations taken on } \hat{F}_{\bm D|\bm x_{t_m}}, \hat{F}_{\bm L|\bm x_{t_m}},
\end{equation*}
which can be efficiently solved by the binary search, because the left-hand side of Equation~\eqref{eq:pb} is increasing in $q_m$ and the right-hand side is decreasing in $q_m$.

The balancing coefficient in Equation~\eqref{eq:balancing_coeff} is used for worst-case performance guarantee. To achieve a better average performance, \citet{chao2015approximation,chao2018approximation} suggested to tune the balancing coefficient $\tilde\beta\in[\beta-0.5,\beta+0.5]$ by simulating the in-sample dynamics and find the best $\tilde\beta$ achieving the lowest in-sample costs. For a more competitive benchmark, we also consider such PTO-PPB (Parameterized PB) policy, which is a boosted version of the PTO-PB policy. The performance of the PTO-PPB benchmark is also included in Appendix~\ref{ecsec:result_PPB}.

\section{Descriptive Analysis for the Real-World Data} \label{ecsec:descriptive}

We conduct a comprehensive descriptive analysis for the real-world data, including the time-series pattern and product-level pattern of the demand and lead time.

First, we show the patterns of the demand data. As shown in Figure~\ref{fig:real_demand_seq}, the demand data exhibits strong periodicity with inherent fluctuations. As shown in Figure~\ref{fig:real_demand_dist}, the average demand distribution across different products is highly skewed: most products experience low demand, while a few exhibit significantly high demand. Since we train a unified model for all products, our inventory management approach naturally prioritizes high-demand products. Consequently, during the training process, we preserved this skewness and did not apply logarithmic transformations to the demand label data.

\begin{figure}[htbp]
     \FIGURE
    {\includegraphics[width=0.6\linewidth]{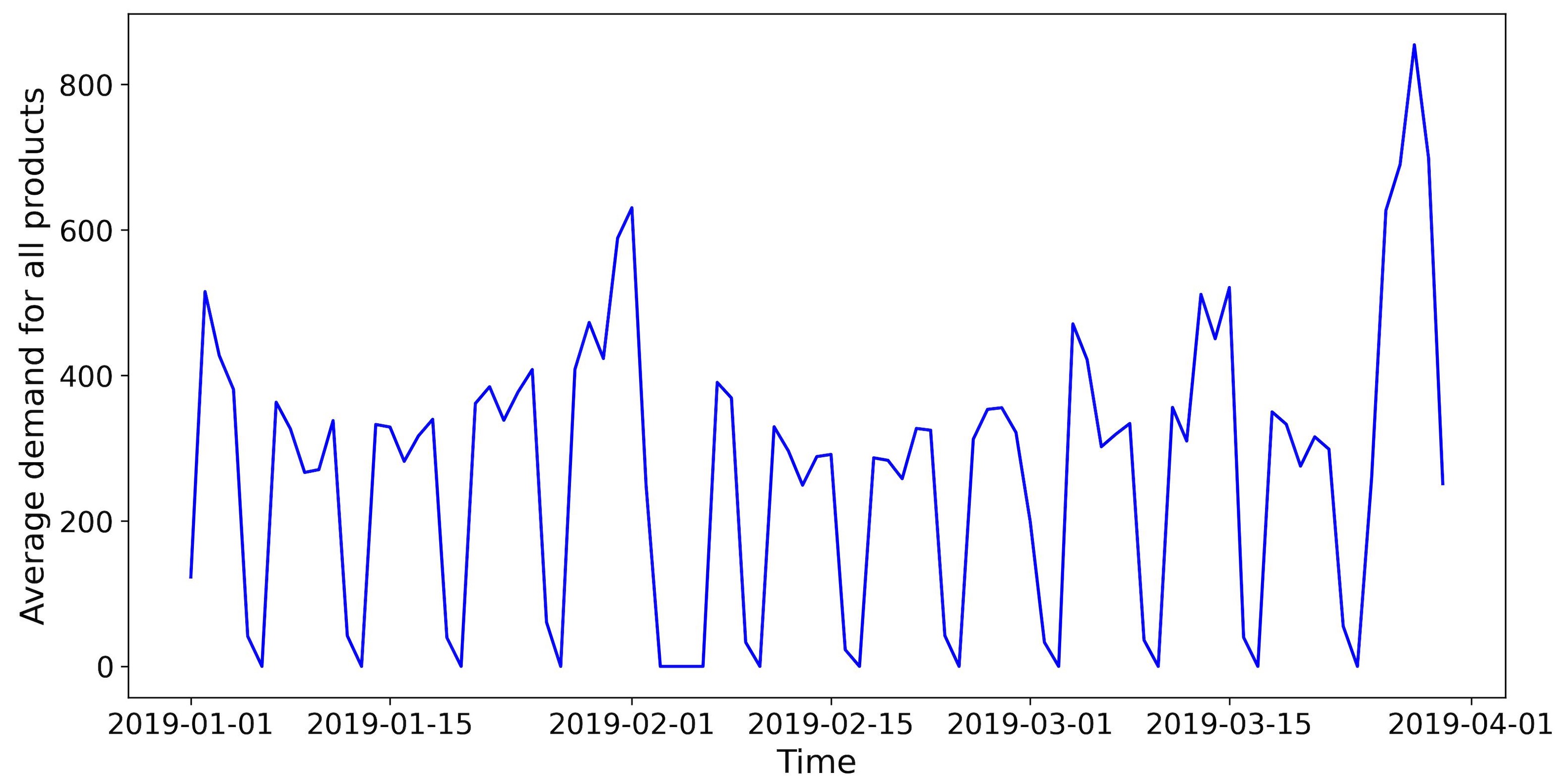}} 
{Average demand sequence from January 2019 to March 2019. The average is computed across all products for each day. \label{fig:real_demand_seq}}
{}
\end{figure}

\begin{figure}[htbp]
     \FIGURE
    {\includegraphics[width=0.7\linewidth]{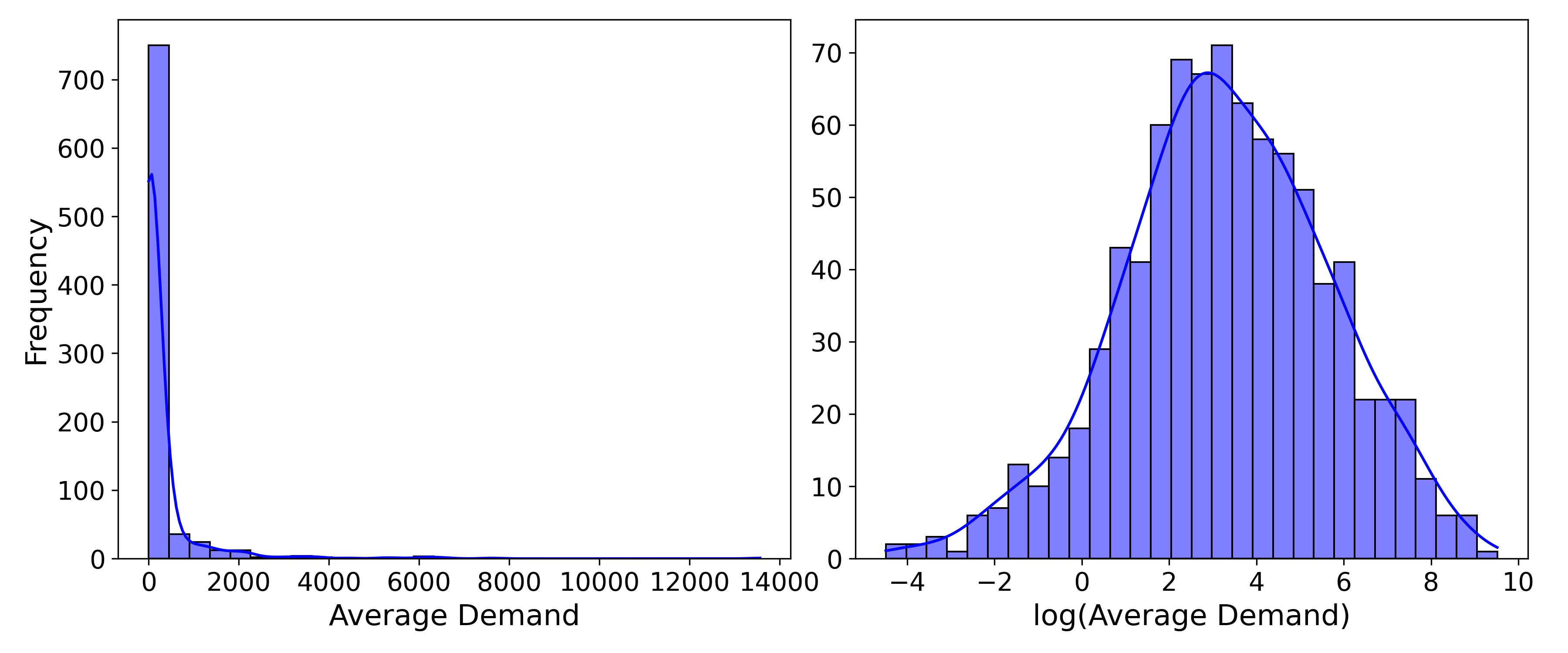}} 
{Histogram of the average demand distribution across products. \label{fig:real_demand_dist}}
{The average demand is computed over the three-month horizon for each product. The left panel shows the raw average demand, while the right panel presents its logarithmic transformation.}
\end{figure}

Second, we show the patterns of the lead time data. As shown in Figure~\ref{fig:leadtime_analysis}, the random lead time ranges from $L=1$ to $9$, with a mean between $3$ and $4$. We analyze the autocorrelation of the sequence of lead times for each product and find that the correlation is generally weak.

\begin{figure}[htbp]
     \FIGURE
    {\includegraphics[width=0.8\linewidth]{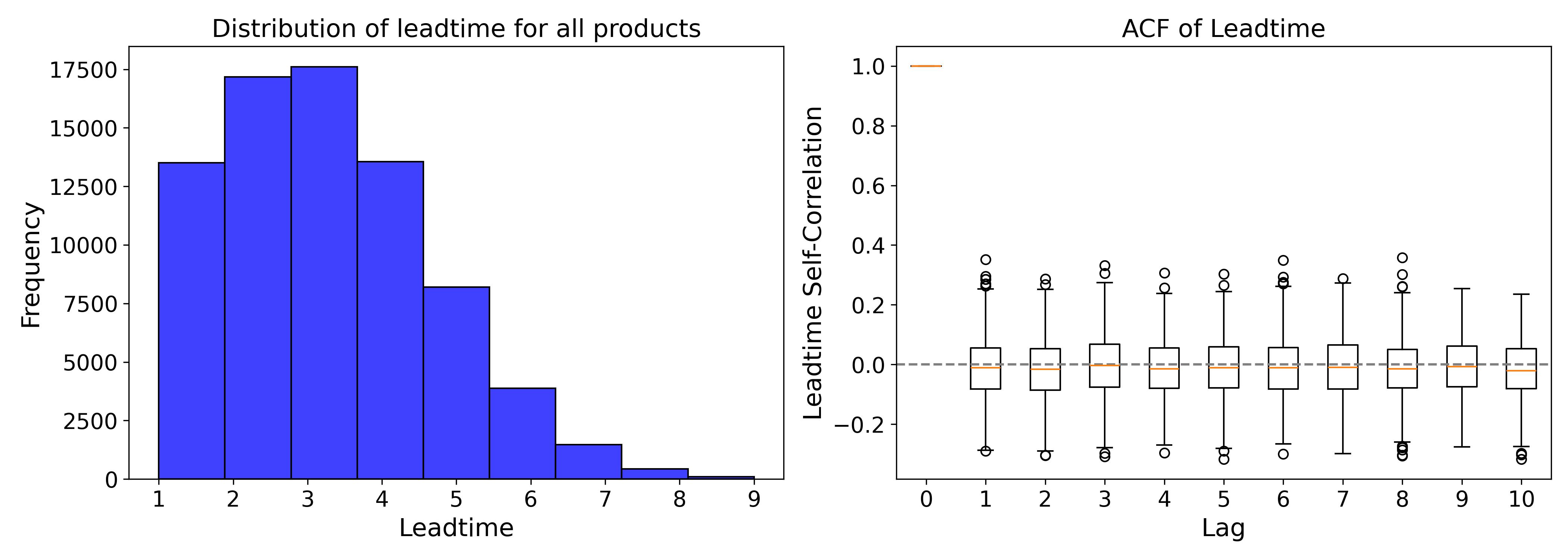}} 
{Lead time analysis. \label{fig:leadtime_analysis}}
{The left panel shows the lead time distribution across all products over the three-month horizon. The right panel presents the autocorrelation function (ACF) for different lags of the sequence of lead times, with the boxplot summarizing results across all products.}
\end{figure}

Furthermore, we examine the relationship between demand and lead time. As illustrated in Figure~\ref{fig:dl_corr}, for most products, there is no significant correlation between demand and lead time, while a few products exhibit weak positive or negative correlations.

\begin{figure}[htbp]
     \FIGURE
    {\includegraphics[width=0.4\linewidth]{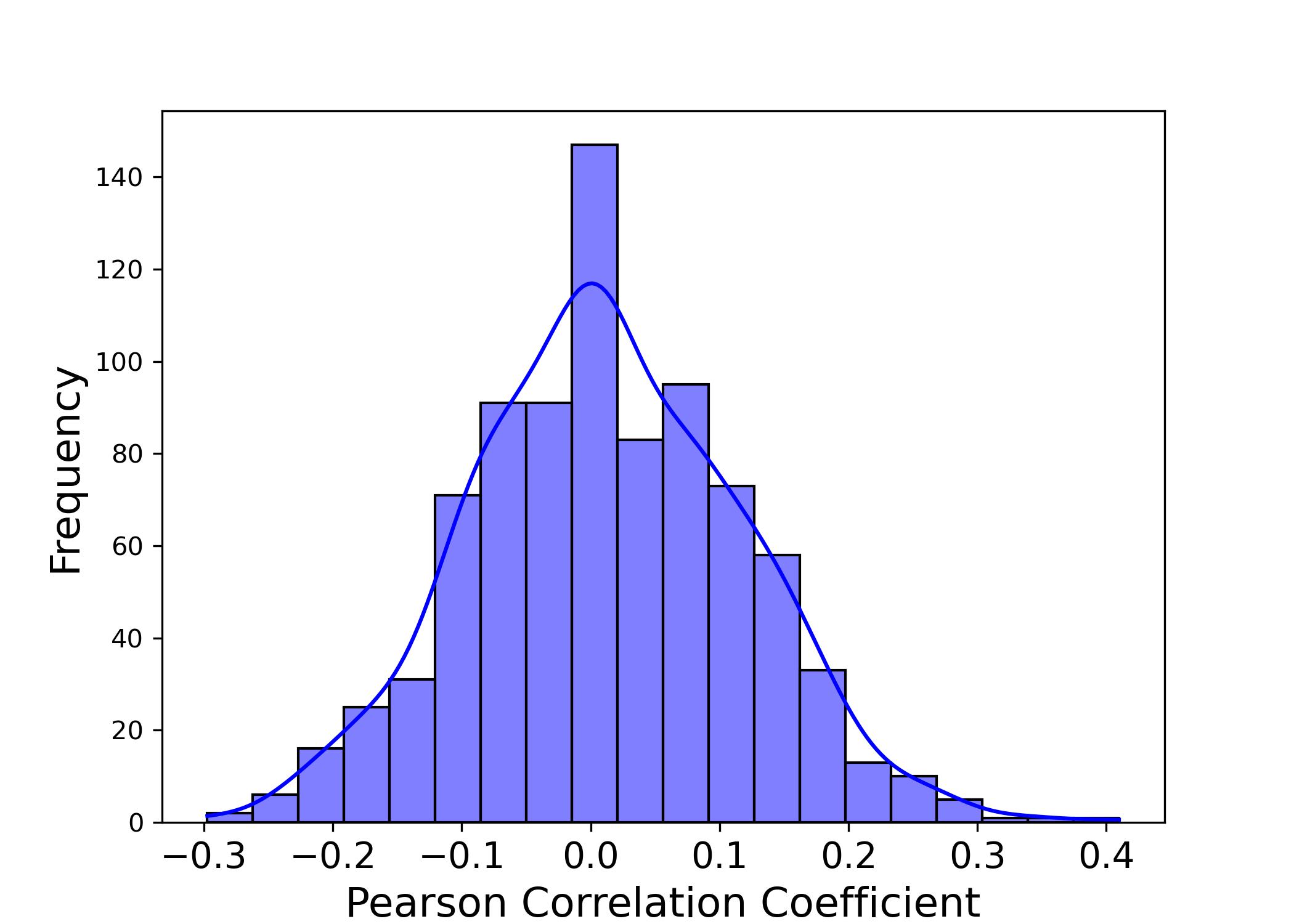}} 
{Distribution of the Pearson correlation coefficient between demand and lead time for each product. \label{fig:dl_corr}}
{}
\end{figure}

\section{Procedure and Results of Hyperparameter Tuning} 
\label{ecsec:tuning}

We first describe the tuning procedure. In our paper, we need to tune two types of hyperparameters. The first type relates to the LSTM model, including key parameters such as the network architecture and learning rate. We call this typer of hyperparameter as ``Network-based'' hyperparameter.  The Network-based hyperparameters are required for both the PTO and E2E models. The hyperparameter search is conducted using the Bayesian searching rule provided by the Optuna Python package. Additionally, we tune these parameters in only one experimental setting, as the resulting optimal hyperparameters are found to be robust and perform well across all settings. Further manual tuning for each specific case does not lead to significant performance gains, while it would incur substantial computational cost. Therefore, we fix the Network-based hyperparameters for all problem settings. The search ranges and best combinations of Network-based hyperparameter in real data and synthetic data for are detailed in Tables~\ref{tab:hyperparams-real} and \ref{tab:hyperparams-syn}.

\begin{table}[htbp!]
\TABLE
{The tuning ranges and best combinations of the Network-based hyperparameters for the PTO and E2E frameworks using the real-world datasets.
    \label{tab:hyperparams-real}}
{\begin{tabular}{l 
                    >{\centering\arraybackslash}m{3.5cm} 
                    >{\centering\arraybackslash}m{2cm} 
                    >{\centering\arraybackslash}m{2cm} 
                    >{\centering\arraybackslash}m{2cm}}
        \toprule
        \multirow{2}{*}{\textbf{Hyperparameter}} & \multirow{2}{*}{\textbf{Range}} & \multicolumn{2}{c}{\textbf{PTO}} & \multirow{2}{*}{\textbf{E2E}} \\
        \cmidrule(lr){3-4}
        & & \textbf{Demand} & \textbf{Lead time} & \\
        \midrule
        Hidden size for demand module        & [32, 64, 128]                         & 128 & –   & 64  \\
        Hidden size for lead time module     & [32, 64, 128]                         & –   & 64  & 32  \\
        Hidden size for order/PIL module & [128, 256]                                     & –   & –   & 256 \\
        Embedding size                       & [0, 1, 5, 10, 15]                     & 5   & 1   & 15  \\
        Number of LSTM layers                   & [1, 2, 3]                             & 3   & 2   & 2   \\
        Weight decay                         & [0, $10^{-6}$, $10^{-5}$, $10^{-4}$]   & $0$ & $10^{-4}$ & $10^{-6}$ \\
        Batch size                           & [64, 128, 256]                        & 256 & 256 & 128 \\
        Learning rate                        & [$10^{-4}$, $10^{-3}$, $10^{-2}$]     & $10^{-2}$ & $10^{-2}$ & $10^{-3}$ \\
        Learning rate decay rate             & [0.4, 0.6, 0.8]                       & 0.6 & 0.8 & 0.8 \\
        Learning rate decay step             & [1, 3, 5]                             & 5   & 1   & 5   \\
        \bottomrule
    \end{tabular}}
{}
\end{table}

\begin{table}[htbp!]
\TABLE
{Selected Network-based hyperparameters for deep learning models under the PTO and E2E frameworks for the synthetic datasets. The search ranges are same as in Table~\ref{tab:hyperparams-real}.
    \label{tab:hyperparams-syn}}
{\begin{tabular}{l 
                    >{\centering\arraybackslash}m{2.2cm} 
                    >{\centering\arraybackslash}m{2.2cm} 
                    >{\centering\arraybackslash}m{2.2cm}}
        \toprule
        \multirow{2}{*}{\textbf{Hyperparameter}} 
            & \multicolumn{2}{c}{\textbf{PTO}} 
            & \multirow{2}{*}{\textbf{E2E}} \\
        \cmidrule(lr){2-3}
            & \textbf{Demand} & \textbf{Lead time} & \\
        \midrule
        Hidden size for demand module        & 128 & –   & 64  \\
        Hidden size for lead time module     & –   & 128 & 64  \\
        Hidden size for order/PIL module & –   & –   & 128 \\
        Embedding size                       & 1   & 1   & 1   \\
        Number of LSTM layers                   & 2   & 2   & 2   \\
        Weight decay                         & $10^{-5}$ & $10^{-4}$ & $10^{-5}$ \\
        Batch size                           & 128 & 128 & 128 \\
        Learning rate                        & $10^{-3}$ & $10^{-4}$ & $10^{-3}$ \\
        Learning rate decay rate             & 0.4 & 0.8 & 0.6 \\
        Learning rate decay step             & 5   & 3   & 5   \\
        \bottomrule
    \end{tabular}}
{}
\end{table}

\begin{table}[htbp!]
\TABLE
{Ranges of the fine-tuned Loss-basesd parameters.
    \label{tab:hyperparams-finetune}}
{\begin{tabular}{l 
                        >{\centering\arraybackslash}m{5cm}}
            \toprule
            \multirow{2}{*}{\textbf{Hyperparameter}} & \multirow{2}{*}{\textbf{Range}} \\
            & \\
            \midrule
            Demand prediction regularization $\lambda^D$ for E2E models 
            & [0, 0.1, 1.0, 2.5] \\
            Lead time prediction regularization $\lambda^L$ for E2E models 
            & [0, 0.01, 0.1, 0.25] \\
            POI prediction regularization $\lambda^{POI}_1$ for E2E models 
            & [0, 0.1, 1.0, 2.5] \\
            POI prediction regularization $\lambda^{POI}_2$ for E2E models 
            & [0, 0.1, 0.5, 1.0] \\
            Balancing coefficient $\tilde{\beta}$ for PTO-PPB policy
            & $\beta + [-0.5, -0.45, \dots, 0.5]$ \\
            Boosting parameter $\gamma$ for E2E-BPIL policy
            & [0.8, 0.85, \dots, 1.4] \\
            \bottomrule
        \end{tabular}}
{\textit{Note.} $\beta$ is the balancing coefficient of the PTO-PB policy.}
\end{table}

The second type of hyperparameters pertains to the loss function, which directly influences the direction of the training objective. For example, this includes the regularization parameters $\lambda^D$, $\lambda^L$, $\lambda^{POI}_1$, and $\lambda^{POI}_2$ in Equation~\eqref{eq:loss}, the balancing coefficient $\tilde{\beta}$ for the PTO-PPB policy, and the boosting parameter $\gamma$ for the E2E-BPIL policy. We call this type of hyperparamter as ``Loss-based" hyperparamter. Since the Loss-based hyperparamters are highly dependent on the experimental setting, we fine-tune them for each setting to ensure that the results are both reasonable and comparable. The parameter ranges used for fine-tuning are listed in Table~\ref{tab:hyperparams-finetune}. Note that a similar procedure which fixed basic hyperparameters and tuning the most critical ones, has also been adopted in \citet{gijsbrechts2022can}.

\section{Synthetic Data Generation} \label{ecsec:dgp}

We consider demand and lead time data generated under four distinct configurations. For each configuration, we generate $20$ instances by sampling the mean $\mu_k$ of the demand feature $x_{t,k}$ from a uniform distribution $U[0,1]$ for $k = 1,2,3,4$ across all products. Using these sampled means, we generate a corresponding sample dataset of size $1000 \times 300$ containing demand and lead time data for all products in each instance. The average performance of a given policy across all instances is computed to ensure robustness. The four configurations are detailly introduced as follows:
\begin{enumerate}
    \item {\emph{Independent and identically distributed demand, Constant lead time (IC):}}  In this configuration, we consider independent and identically distributed (i.i.d.) demand with a constant lead time, abbreviated as IC. Specifically, all demand features and noise terms are i.i.d. normal variables across the time horizon, i.e., $x_{t,k} \sim N(\mu_k, (\sigma_k)^2)$. The standard deviation is set as $\sigma_k = 0.6\mu_k$, and the noise term $\epsilon_t$ follows a standard normal distribution, i.e., $\epsilon_t \sim N(0,1)$. The lead time remains constant at $L_t \equiv 3$. 
    \item {\emph{Correlated demand, Constant lead time (CC):}} In this configuration, we introduce time-correlated demand while keeping the lead time constant, abbreviated as CC. Specifically, both the demand features and the noise follow an autoregressive process (AR(1)). The demand features evolve according to  $x_{t+1,k} = 0.2\mu_k + 0.8x_{t,k} + \eta_{t,k}$, $\eta_{t,k} \sim N(0, (0.36\mu_k)^2)$. And the noise term evolves as  $\epsilon_{t+1} = 0.8\epsilon_t + \xi_t$, $\xi_t \sim N(0, 0.36)$. The lead time remains constant at $L_t \equiv 3$. 

    \item {\emph{Correlated demand, Random correlated lead time (CR):}} In this configuration, we retain the demand generation process from the CC setting but introduces a time-varying, random lead time, abbreviated as CR. The lead time is modeled as an AR(1) process: $L_{t+1} = 0.6 + 0.8L_t + \gamma_t$, where $\gamma_t \sim N(0,1)$.

    \item {\emph{Shock on Correlated demand, Random correlated lead time (SCR)}}: In this configuration, we incorporate a common shock to correlate demand with random lead time, based on the CR setting. This configuration is abbreviated as SCR. The demand features follow the same process as in the CR setting, while the noise term is decomposed into two components: $\epsilon_t = \tilde\epsilon_t + \sqrt{0.27}s_t$, where $\tilde\epsilon_{t+1} = 0.8\tilde\epsilon_t + N(0, 0.09)$, and $s_{t+1} = 0.8s_t + N(0,1)$. The common shock $s_t$ also influences the lead time, which is modeled as $L_t = \tilde L_t + \sqrt{0.75}s_t$, where $\tilde L_{t+1} = 0.6 + 0.8\tilde L_t + N(0,0.25)$. 
\end{enumerate}

Across all four configurations, we maintain consistent long-term means and variances for the demand features, noise, and lead time, except for the variations introduced by the constant or stochastic nature of the lead time. Notably, as we transition from the IC to the SCR configuration, the underlying processes become progressively more complex. To ensure data stationarity, we generate 360 days of data and use only days $31$ to $330$ as the final dataset.

\section{More Numerical Results} \label{ecsec:more_results}

\subsection{The Performance of the PTO-PPB Policy} \label{ecsec:result_PPB}

We further compare the more competitive PTO-PPB benchmark with our E2E-PIL and E2E-BPIL policies. For the real-world data in Figure~\ref{fig:real_sensitivity_PPB}, the PTO-PPB policy performs worse than PTO-PB in most cases, which is likely caused by overfitting during training and tuning, as the balancing coefficient in PTO-PB may already be empirically near-optimal for this dataset. In contrast, for the synthetic data in Figure~\ref{fig:syn_sensitivity_PPB}, the PTO-PPB policy outperforms PTO-PB, whereas the E2E-PIL policy still achieves lower costs in most configurations except when the unit backorder cost coefficient is high. Under these conditions, the boosted E2E-BPIL policy further corrects the order quantity and ultimately attains the best performance.

\begin{figure}[htbp]
     \FIGURE
    {\includegraphics[width=0.9\linewidth]{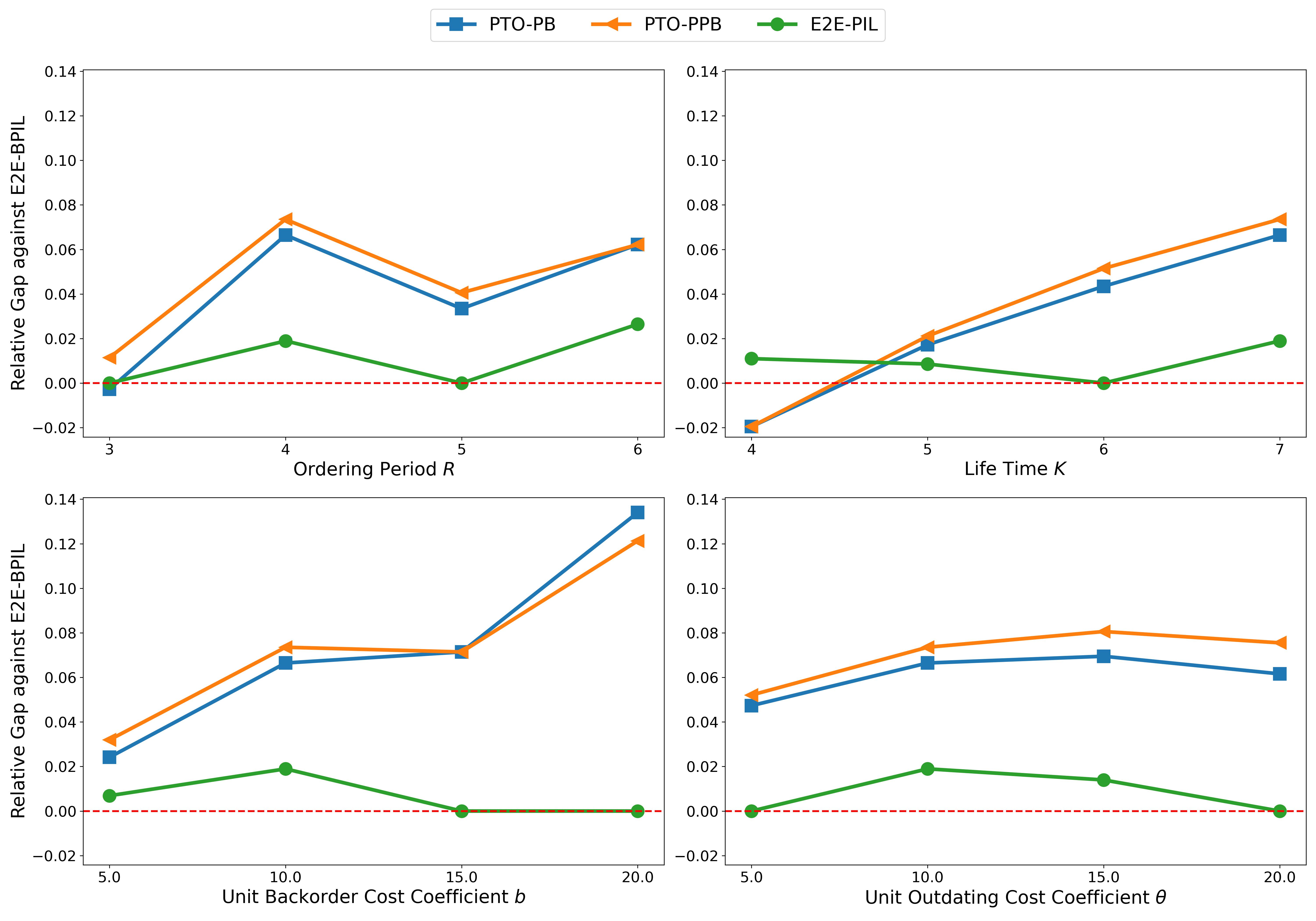}} 
{Relative gap of PTO-(P)PB and E2E-PIL against E2E-BPIL in real-world data across different problem settings. \label{fig:real_sensitivity_PPB}}
{}
\end{figure}

\begin{figure}[htbp]
     \FIGURE
    {\includegraphics[width=0.9\linewidth]{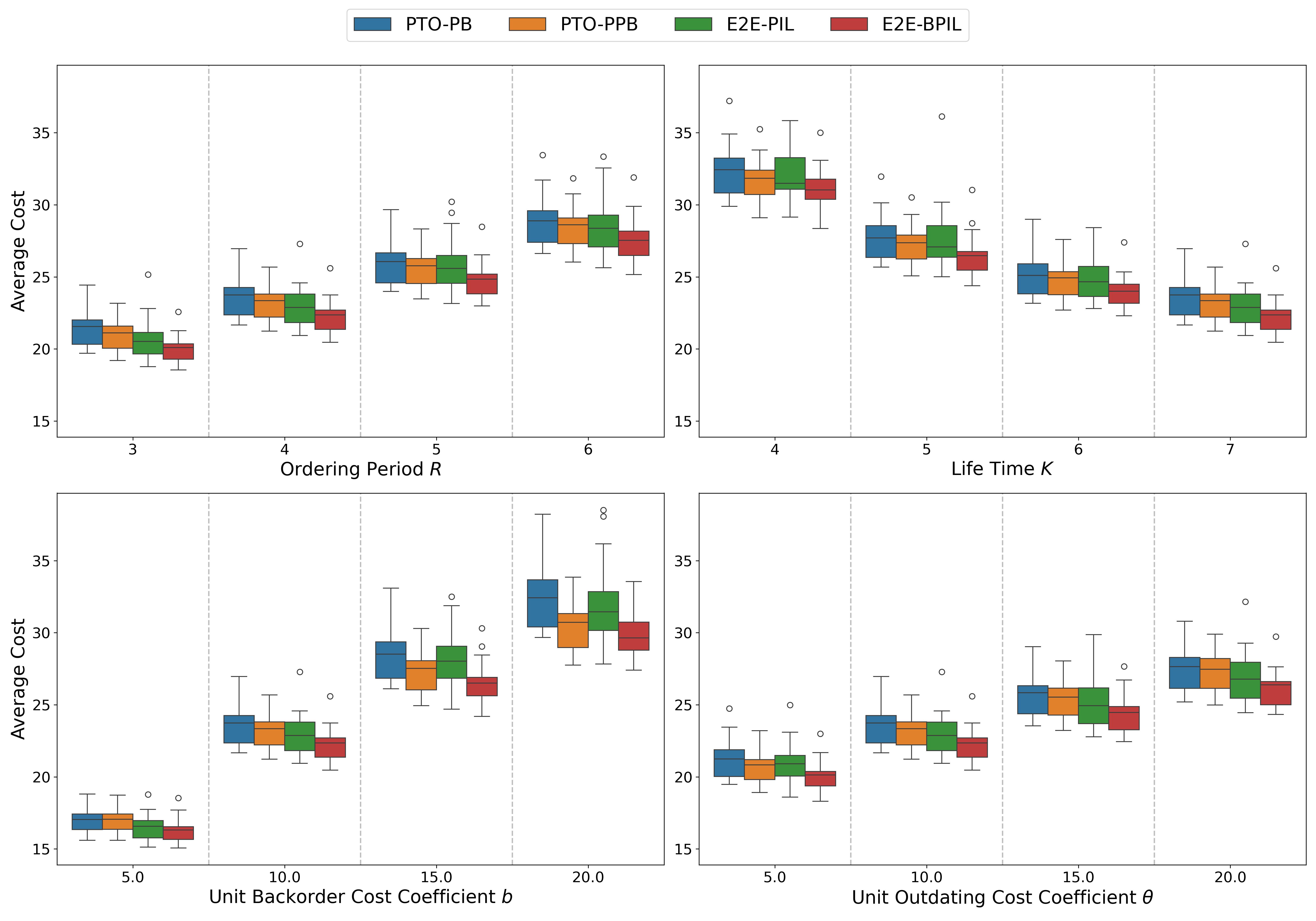}} 
{Sensitivity boxplot of average cost of PTO-(P)PB and E2E-(B)PIL policies in synthetic data (SCR setting) across different problem settings.\label{fig:syn_sensitivity_PPB}}
{}
\end{figure}

\subsection{T-tests to Compare the Performance of the PTO and E2E Policies} \label{ecsec:t_test}

Under the baseline problem setting, we compare two PTO policies (PTO-PB and PTO-PPB) with three E2E policies (E2E-BB, E2E-PIL, and E2E-BPIL), and conduct pairwise two-sample t-tests using 20 instances as independent observations. The null hypothesis is $H_0: C^{\mathrm{PTO}}\le C^{\mathrm{E2E}}$, so larger t-statistics and smaller p-values provide stronger evidence against $H_0$, supporting the conclusion that an E2E policy achieves a significantly lower average cost than its PTO counterpart. The test results are summarized in Table~\ref{tab:t_test_pto_vs_e2e}. 

E2E-PIL significantly outperforms PTO-PB in the IC and CC settings and yields comparable t-values in the CR and SCR settings. Meanwhile, the boosted E2E-BPIL policy consistently outperforms the boosted PTO-PPB policy. These results collectively highlight the advantage of the theory-guided E2E design in both the original and boosted comparisons.

\begin{table}[htbp!]
\TABLE
{T-statistics (with p-value in parentheses) for comparisons between PTO and E2E policies.
    \label{tab:t_test_pto_vs_e2e}}
{ \begin{tabular}{lcccc}
      \toprule
      Pair/Configuration & IC & CC & CR & SCR \\
      \midrule
      PTO-PB vs E2E-BB & 2.206(0.017)* & -0.623(0.732) & -1.106(0.862) & -2.162(0.982) \\
      PTO-PB vs E2E-PIL & 2.551(0.007)** & 2.341(0.012)* & 0.575(0.284) & 1.362(0.091) \\
      PTO-PB vs E2E-BPIL & 3.062(0.002)** & 3.317(0.001)** & 2.268(0.015)* & 3.413(0.001)** \\
      PTO-PPB vs E2E-BB & 1.400(0.085) & -0.831(0.794) & -1.109(0.863) & -2.887(0.997) \\
      PTO-PPB vs E2E-PIL & 1.740(0.045)* & 2.168(0.018)* & 0.579(0.283) & 0.570(0.286) \\
      PTO-PPB vs E2E-BPIL & 2.253(0.015)* & 3.185(0.001)** & 2.292(0.014)* & 2.689(0.005)** \\
      \bottomrule
    \end{tabular}}
{\textit{Note.} Significance levels: $^{***}p<0.001$, $^{**}p<0.01$, $^{*}p<0.05$.}
\end{table}

\subsection{Full Results of the Synthetic Data in the Baseline Setting} \label{ecsec:full_result_syn}

To provide more numerical details, we record the holding costs, backorder and outing costs per period, as well as the stockout rate (i.e., the rate of periods with positive unmet demands) and the outdating rate (i.e., the rate of periods with positive inventory perishments) of the baseline problem setting. The full results are shown in Table~\ref{tab:ic_measures}-\ref{tab:crc_measures}.

For the trade-off between inventory overage and underage, E2E-PIL typically achieves a more balanced inventory position than the competing policies. Relative to the PTO-(P)PB policies, E2E-PIL tends to order less aggressively, leading to lower holding and outdating costs, along with a lower outdating rate. Relative to E2E-BB, it is less prone to under-ordering and thus tends to incur lower backorder costs and a lower stockout rate. This improved balance between overage and underage often translates into a lower total cost. Moreover, by leveraging the constant boosting technique of the ODA framework, the E2E-BPIL slightly increases inventory levels to substantially reduce backorders, while inducing only a modest increase in holding and outdating costs, thereby further improving overall performance.

\begin{table}[htbp!]
\TABLE
{Average performance in the IC (i.i.d demand and constant lead time) setting, with standard deviation in the parentheses.
    \label{tab:ic_measures}}
{ \begin{tabular}{lcccccc}
    \toprule
    Policy/Measure & Total Cost & Holding Cost & Backorder Cost & Outdating Cost & Stockout Rate & Outdating Rate \\
    \midrule
    PTO-PB   & 9.317(0.525) & 5.093(0.281) & 3.787(0.494) & 0.437(0.074) & 0.144(0.011) & 0.024(0.004) \\
    PTO-PPB  & 9.167(0.479) & 5.737(0.328) & 2.830(0.411) & 0.600(0.089) & 0.112(0.010) & 0.031(0.004) \\
    E2E-BB   & 8.934(0.545) & 5.148(0.529) & 3.394(0.744) & 0.392(0.105) & 0.133(0.024) & 0.023(0.005) \\
    E2E-PIL  & 8.886(0.517) & 5.256(0.422) & 3.201(0.556) & 0.429(0.085) & 0.126(0.016) & 0.024(0.004) \\
    E2E-BPIL & 8.814(0.487) & 5.602(0.370) & 2.703(0.294) & 0.508(0.085) & 0.109(0.008) & 0.028(0.004) \\
    \bottomrule
  \end{tabular}}
{}
\end{table}

\begin{table}[htbp!]
\TABLE
{Average performance in the CC (time-correlated demand and constant lead time) setting, with standard deviation in the parentheses.
    \label{tab:cc_measures}}
{\begin{tabular}{lcccccc}
    \toprule
    Policy/Measure & Total Cost & Holding Cost & Backorder Cost & Outdating Cost & Stockout Rate & Outdating Rate \\
    \midrule
    PTO-PB   & 19.480(1.190) & 7.607(0.486) & 8.357(1.392) & 3.515(0.386) & 0.155(0.016) & 0.082(0.007) \\
    PTO-PPB  & 19.480(1.161) & 7.868(0.520) & 7.913(1.427) & 3.699(0.415) & 0.147(0.017) & 0.085(0.007) \\
    E2E-BB   & 20.237(2.738) & 5.574(0.957) & 12.486(3.821) & 2.178(0.530) & 0.224(0.052) & 0.060(0.011) \\
    E2E-PIL  & 19.146(2.232) & 6.183(0.697) & 10.406(3.114) & 2.557(0.456) & 0.194(0.038) & 0.065(0.009) \\
    E2E-BPIL & 18.539(1.361) & 7.391(0.552) & 7.822(1.734) & 3.326(0.403) & 0.152(0.021) & 0.077(0.007) \\
    \bottomrule
  \end{tabular}}
{}
\end{table}

\begin{table}[htbp!]
\TABLE
{Average performance in the CR (time-correlated demand and random lead time) setting, with standard deviation in the parentheses.
    \label{tab:cr_measures}}
{ \begin{tabular}{lcccccc}
    \toprule
    Policy/Measure & Total Cost & Holding Cost & Backorder Cost & Outdating Cost & Stockout Rate & Outdating Rate \\
    \midrule
    PTO-PB   & 15.987(1.089) & 6.427(0.413) & 6.973(1.326) & 2.588(0.293) & 0.155(0.018) & 0.066(0.006) \\
    PTO-PPB  & 15.885(0.979) & 7.048(0.441) & 5.885(1.172) & 2.952(0.322) & 0.132(0.017) & 0.072(0.006) \\
    E2E-BB   & 16.323(2.077) & 4.909(0.730) & 9.698(2.905) & 1.716(0.365) & 0.215(0.047) & 0.05(0.008) \\
    E2E-PIL  & 15.055(1.352) & 5.887(0.585) & 6.886(1.873) & 2.281(0.405) & 0.167(0.031) & 0.059(0.008) \\
    E2E-BPIL & 14.834(1.054) & 6.616(0.507) & 5.534(1.449) & 2.683(0.354) & 0.138(0.023) & 0.066(0.007) \\
    \bottomrule
  \end{tabular}}
{}
\end{table}

\begin{table}[htbp!]
\TABLE
{Average performance in the SCR (time-correlated demand, random lead time, and demand and lead time are correlated by a common shock) setting, with standard deviation in the parentheses.
    \label{tab:crc_measures}}
{ \begin{tabular}{lcccccc}
    \toprule
    Policy/Measure & Total Cost & Holding Cost & Backorder Cost & Outdating Cost & Stockout Rate & Outdating Rate \\
    \midrule
    PTO-PB   & 23.607(1.318) & 7.540(0.466) & 12.304(1.574) & 3.764(0.353) & 0.195(0.015) & 0.085(0.006) \\
    PTO-PPB  & 23.235(1.137) & 8.891(0.512) & 9.553(1.336) & 4.792(0.425) & 0.156(0.013) & 0.099(0.006) \\
    E2E-BB   & 24.891(2.227) & 5.841(0.875) & 16.453(3.423) & 2.597(0.572) & 0.251(0.043) & 0.067(0.010) \\
    E2E-PIL  & 22.993(1.460) & 6.749(0.572) & 13.031(2.021) & 3.213(0.449) & 0.210(0.020) & 0.074(0.007) \\
    E2E-BPIL & 22.234(1.158) & 8.408(0.450) & 9.410(1.378) & 4.416(0.381) & 0.159(0.014) & 0.091(0.006) \\
    \bottomrule
  \end{tabular}}
{}
\end{table}

\section{Proofs}
\label{ecsec:proofs}

\subsection{Proof of Proposition~\ref{prop_oda}}

\proof{Proof.}  We first show, by induction, that all inventory state vectors $\bm z_{t}$ are scaled by $\gamma$ for all $t=1,\ldots,T$. We assume the initial inventory state to be $\bm z_1 \equiv \bm 0 = \gamma \bm 0$, which is already scaled by $\gamma$. For a general initial state, one can simply scale it by $\gamma$ in the same manner, thereby ensuring consistency with the scaling applied to the demand and decision sequences. Assume that at time $t$, the state $\bm z_t$ is scaled by $\gamma$. If time $t$ is not an ordering point, then according to the system transitions in Equation~\ref{eq:transit}, we have
\begin{subequations}  \label{eq:scaled_transit}
    \begin{align*}
        z^\pi_{t+1,i} &=(\gamma z^\pi_{t,i+1}-(\gamma D_{t}-\sum_{j=1}^{i}\gamma z^\pi_{t,j})^{+})^{+}=\gamma\left[( z^\pi_{t,i+1}-(D_{t}-\sum_{j=1}^{i} z^\pi_{t,j})^{+})^{+}\right], &\mathrm{for}\ i=1,\ldots,K-1;  \\
        z^\pi_{t+1,i} &=\gamma z^\pi_{t,i+1}-(\gamma D_t-\sum_{j=1}^i \gamma z^\pi_{t,j})^+=\gamma\left[z^\pi_{t,i+1}-(D_t-\sum_{j=1}^i z^\pi_{t,j})^+\right],&\mathrm{for}\ i=K;  \\
        z^\pi_{t+1,i} &=\gamma z^\pi_{t,i+1},&\mathrm{for}\ i=K+1,\ldots,K+\bar{L}-2; \\
        z^\pi_{t+1,i} &=0=\gamma\cdot 0,&\mathrm{for}\ i=K+\bar{L}-1.
    \end{align*}
\end{subequations}
If time $t$ is an ordering point (i.e., $\exists m\in{1,\ldots,M}$ such that $t=t_m$), we additionally have
\begin{equation*}
    z^\pi_{t+1,K+L_m-1}=\gamma z^\pi_{t+1,K+L_m-1}+\gamma q^\pi_m = \gamma(z^\pi_{t+1,K+L_m-1}+ q^\pi_m).
\end{equation*}
Hence, in either case, the next inventory state $\bm z_{t+1}$ is also scaled by $\gamma$. By induction, all inventory states satisfy this homogeneous property.

Finally, the total cost satisfies
\begin{align*}
    C(\{\gamma D_t\}_{t=1}^T,\{\gamma q^\pi_m\}_{m=1}^M)&=\sum_{t=1}^{T}\left (h(\sum_{i=1}^K\gamma z^\pi_{t,i}-\gamma D_t)^++b(\gamma D_t-\sum_{i=1}^K\gamma z^\pi_{t,i})^++\theta(\gamma z^\pi_{t,1}-\gamma D_t)^+\right )\\
    &=\gamma \sum_{t=1}^{T}\left (h(\sum_{i=1}^K z^\pi_{t,i}-D_t)^++b( D_t-\sum_{i=1}^K z^\pi_{t,i})^++\theta( z^\pi_{t,1}- D_t)^+\right )\\
    &=\gamma C(\{D_t\}_{t=1}^T,\{ q^\pi_m\}_{m=1}^M),
\end{align*}
which completes the proof. \Halmos

\subsection{Proof of Lemma~\ref{lem:excess_risk}}

\proof{Proof.}  Let $f^*\in \mathop{\arg\min}_{f\in\mathcal F}\ \mathcal R(f)$ be the population risk minimizer within the hypothesis space $\mathcal F$. 

Then we have
\begin{align*}
    \Delta \mathcal{R}(\hat{f}_n)&= \mathcal R(\hat{f}_n)-\mathcal R(g^*)\\
    &= \big[\mathcal R(\hat{f}_n)-\hat{\mathcal R}_n(\hat{f}_n)\big]+\big[\hat{\mathcal R}_n(\hat{f}_n)-\mathcal R(f^*)\big]+\big[\mathcal R(f^*)-\mathcal R(g^*)\big]\\
    &\leq \big[\mathcal R(\hat{f}_n)-\hat{\mathcal R}_n(\hat{f}_n)\big]+\big[\hat{\mathcal R}_n(f^*)-\mathcal R(f^*)\big]+\big[\mathcal R(f^*)-\mathcal R(g^*)\big]\\
    &\leq\big|\mathcal R(\hat{f}_n)-\hat{\mathcal R}_n(\hat{f}_n)\big|+\big|\mathcal R(f^*)-\hat{\mathcal R}_n(f^*)\big|+\big[\mathcal R(f^*)-\mathcal R(g^*)\big]\\
    &\leq2\sup_{f\in\mathcal F}\big|\mathcal R(f)-\hat{\mathcal R}_n(f)\big|+\inf_{f\in \mathcal F}\big[\mathcal R(f)-\mathcal R(g^*)\big].
\end{align*}

The third row is due to the fact that $\hat{f}_n$ is the empirical risk minimizer and hence $\hat{\mathcal R}_n(\hat{f}_n)\leq\hat{\mathcal R}_n(f^*)$.\Halmos
\endproof

\subsection{Proof of Lemma~\ref{lem:poly_dnn}}

\proof{Proof.}  
We first show that $\mathcal F_{\mathcal A}^{\text{poly}}$ satisfies Assumption~\ref{asp:hypothesis} for any $\mathcal A\geq 1$. For any $f\in \mathcal F_{\mathcal A}^{\text{poly}}$, its ``fix-and-shift'' transform for any constant $c\in\mathbb R$ is
$$
f_c(\bm x,z)=f(\bm x,c)+c-z=\sum_{0\leq i_1+\cdots+i_p+j\leq \mathcal A}\left( \alpha _{i_1,\dots,i_{p+1}}c^j\Pi_{k=1}^px_{k}^{i_k}\right)+c-z.
$$
Rearranging by monomial degrees in $(x_1,\dots,x_p)$ gives
\begin{align*}
    f_c(\bm x,z)=& \left (c+\sum_{j=0}^\mathcal Ac^j\alpha_{0,\dots,0,j}\right)
    +\sum_{i_1+\cdots+i_p=1}\left (\sum_{j=0}^{\mathcal A-1}\alpha_{i_1,\dots, i_p,j}c^{j}\right)\Pi_{k=1}^px_{k}^{i_k}+\cdots\\
    &+\sum_{i_1+\cdots+i_p=\mathcal A-1}\left (\sum_{j=0}^{1}\alpha_{i_1,\dots, i_p,j}c^{j}\right)\Pi_{k=1}^px_{k}^{i_k}
    +\sum_{i_1+\cdots+i_p=\mathcal A}\left (\alpha_{i_1,\dots, i_p,0}\right)\Pi_{k=1}^px_{k}^{i_k} -z.
\end{align*}
All new coefficients are real because they are finite linear combinations of the original $\alpha_{i_1,\dots,i_p,j}$ and powers of $c$. Moreover, the only remaining $z$-term is the linear term $-z$, and coefficients of higher-order powers of $z$ and cross-terms of $\bm x$ and $z$ equal to $0$. Therefore,  $f_c\in{\mathcal F}_{\mathcal A}^{\text{poly}},\forall\,c\in\mathbb R$.

We next show that $\mathcal F_{\mathcal A,\bm w}^{\text{DNN}}$ also satisfies Assumption~\ref{asp:hypothesis} for any $\mathcal A \ge 1$ and $w_i \ge 1$, using a similar re-parameterization argument. For any $f\in \mathcal F_{\mathcal A,\bm w}^{\text{DNN}}$, its ``fix-and-shift'' transform for any constant $c\in\mathbb R$ is
$$
f_c(\bm x,z)=f(\bm x,c)+c-z=l^{\text{aug}}((\bm x;c))+l_\mathcal A\circ \text{ReLU}\circ \cdots \circ \text{ReLU}\circ l_0((\bm x;c))+c-z.
$$

We re-parameterize the network by $\{\bm W^{\text{aug}'},(\bm W_i',\bm b_i')_{i=0}^{\mathcal A}\}$ so that it can represent $f_c(\bm x,z)$ exactly.

\begin{itemize}
    \item \emph{First layer:} For each hidden unit $j$, the first $p$ columns of the weight matrix are kept, the $(p+1)$-th column corresponding to $z$ is set to zero, and the bias absorbs the effect of $z=c$:  
    $$
    (\bm W_0')_{j,1:p} = (\bm W_0)_{j,1:p}, \quad (\bm W_0')_{j,p+1} = 0, \quad (\bm b_0')_j = (\bm b_0)_j + (\bm W_0)_{j,p+1}\cdot c
    $$
    where $(\bm W_0)_{j,1:p}$ denotes the vector consisting of the first $p$ elements of the $j$-th row of $\bm W$, and $(\bm W_0)_{j,p+1}$ denotes the $(j,p+1)$-th element of the matrix.
    
    \item \emph{Intermediate layers:} Copy weights and biases directly:  
    $$
    \bm W_i' = \bm W_i, \quad \bm b_i' = \bm b_i, \quad i=1,\dots,\mathcal A-1.
    $$
    
    \item \emph{Final layer:} Keep the weight matrix, and add $c$ plus the contribution from the linear connection at $z=c$ to the bias:  
    $$
    \bm W_\mathcal A' = \bm W_\mathcal A, \quad \bm b_\mathcal A' = \bm b_\mathcal A + c + (\bm W^{\text{aug}})_{p+1}\cdot c.
    $$
    
    \item \emph{Linear connection layer:} Maintain the original weights for $\bm x$, and set the $z$-coordinate weight to $-1$:  
    $$
    (\bm W^{\text{aug}'})_{1:p} = (\bm W^{\text{aug}})_{1:p}, \quad (\bm W^{\text{aug}'})_{p+1} = -1.
    $$
\end{itemize}

After this re-parameterization, the new network computes $f_c(\bm x,z)$ exactly, where the input $z$ does not propagate through the hidden layers and its contribution evaluated at $z=c$ is absorbed into the biases of the hidden and output layers, while the remaining linear effect of $z$ is preserved in the linear connection layer. Therefore, $f_c\in{\mathcal F}^{\text{DNN}}_{\mathcal A,\bm w},\forall\,c\in\mathbb R$.\Halmos
\endproof

\subsection{Proof of Proposition~\ref{prop:cons_H}}

\proof{Proof.} Obviously we have $\mathcal F' \subseteq \mathcal F$, and thus the generation error of $\mathcal F'$ cannot exceed that of $\mathcal F$. For the approximation error we trivially have
$$
\inf_{f\in \mathcal F}\mathcal R(f)\leq \inf_{f\in \mathcal F'}\mathcal R(f).
$$

We now prove the reverse inequality. Since $\mathcal F$ is compact and the Newsvendor loss is Lipschitz continuous, the minimization is attainable within $\mathcal F$. Denoting
$
f^*\in \arg\min_{f\in\mathcal F}\mathcal R(f)
$, we then have
$$
\inf_{f\in \mathcal F}\mathcal R(f)=\mathcal R(f^*)=\mathbb E_{\bm X,z,D}\!\Big[
b\left(D-(f^*(\bm X,z)+z)\right)^+
+h\left((f^*(\bm X,z)+z)-D\right)^+
\Big].
$$

Because $z$ is determined by past demands while $(\bm X,D)$ is i.i.d.\ across periods, $z$ is independent of the current $(\bm X,D)$. Therefore conditioning on $z$ does not change the distribution of $(\bm X,D)$, and we can rewrite
$$
\mathcal R(f^*)=\mathbb E_{z}\Big[\,\mathbb E_{\bm X,D}\big[b\left(D-(f^*(\bm X,z)+z)\right)^+
+h\left((f^*(\bm X,z)+z)-D\right)^+\big]\Big].
$$

Now choose
$$
c^*\in \arg\min_{z\in\mathbb R}\ \mathbb E_{\bm X,D}\left[b\left(D-(f^*(\bm X,z)+z)\right)^+
+h\left((f^*(\bm X,z)+z)-D\right)^+\right],
$$
which is a constant since the distribution of $(\bm X,D)$ does not change with $z$. We then have
$$
\mathbb E_{\bm X,D}\big[b\left(D-(f^*(\bm X,c^*)+c^*)\right)^+
+h\left((f^*(\bm X,c^*)+c^*)-D\right)^+\big]\leq \mathcal R(f^*).
$$

Now applying the fix-and-shift transformation of $f^*$ with $c^*$, we get the extended function $f^*_{c^*}$ satisfying
$$
f^*_{c^*}(\bm x,z)=f^*(\bm x,c^*)+c^*-z, \, \forall (\bm x,z)\in\mathbb R^p\times\mathbb R.
$$

By Assumption~\ref{asp:hypothesis}, $f^*_{c^*}\in\mathcal F$, and it satisfies
\begin{align*}
    \mathcal R(f^*_{c^*})&=\mathbb E_{z}\Big[\,\mathbb E_{\bm X,D}\big[b\left(D-(f^*_{c^*}(\bm X,z)+z)\right)^+
+h\left((f^*_{c^*}(\bm X,z)+z)-D\right)^+\big]\Big]\\
&=\mathbb E_{z}\Big[\,\mathbb E_{\bm X,D}\big[b\left(D-(f^*(\bm X,c^*)+c^*)\right)^+
+h\left((f^*(\bm X,c^*)+c^*)-D\right)^+\big]\Big]\\
&=\,\mathbb E_{\bm X,D}\Big[b\left(D-(f^*(\bm X,c^*)+c^*)\right)^+
+h\left((f^*(\bm X,c^*)+c^*)-D\right)^+\Big]\\
&\leq\mathcal R(f^*).
\end{align*}

Therefore, $f^*_{c^*}$ is also optimal over $\mathcal F$. Moreover, $f^*_{c^*}$ has the base-stock form, and we denote
$$
v^*(\bm x):=f^*(\bm x,c^*)+c^*, \, \forall \bm x\in\mathbb R^p,
$$
which is $z$-independent. We have $v^*\in\mathcal V$, where $\mathcal V$ is the reduced space defined in Equation~\eqref{eq:reduced_V}. Therefore, $f^*_{c^*}$ also lies in $\mathcal F'$ since $f^*_{c^*}(\bm x,z)=v^*(\bm x)-z$. Therefore,
$$
\inf_{f\in\mathcal F'}\mathcal R(f)\le \mathcal R(f^*_{c^*})\leq\mathcal R(f^*)=\inf_{f\in\mathcal F}\mathcal R(f),
$$
which combined with the trivial inequality in the other direction gives
$$
\inf_{f\in\mathcal F}\mathcal R(f)=\inf_{f\in\mathcal F'}\mathcal R(f).
$$
Thus the approximation errors of $\mathcal F$ and $\mathcal F'$ coincide. \Halmos

\endproof
\end{APPENDIX}

%%%%%%%%%%%%%%%%%
\end{document}